\documentclass{cta-author}

{}
{}
{}
\usepackage{rotating}
\usepackage {graphicx,color, multirow}
\usepackage[hang,center]{subfigure}
\usepackage{url}
\RequirePackage{latexsym,amsmath,amssymb}

\usepackage{mathtools}
\usepackage{amsmath}
\usepackage{amssymb}
\usepackage{amsfonts,bm,multirow}
\usepackage{booktabs}  
\usepackage{makecell} 
\usepackage{bbding} 
\usepackage[switch]{lineno}  
\usepackage{bm}
\usepackage{siunitx}
\newcolumntype{d}[1]{D{.}{.}{#1}}

\begin{document}
	
	\supertitle{Submission Template for IET Research Journal Papers}
	
	\title{Anomaly Detection in Video Sequences: A Benchmark and Computational Model}
	
	\author{\au{Boyang Wan$^{1}$}, \au{Wenhui Jiang$^{1}$}, \au{Yuming Fang$^{1\corr}$}, \au{Zhiyuan Luo$^{1}$}, \au{Guanqun Ding$^{1}$}}
	
	\address{\add{1}{School of Information Management, Jiangxi University of Finance and Economics, 665 Yuping West Street, Qingshanhu District, Nanchang City, Nanchang, China}
		\email{fa0001ng@e.ntu.edu.sg}}
	
	\begin{abstract}
		Anomaly detection has attracted considerable search attention. However, existing anomaly detection databases encounter two major problems. Firstly, they are limited in scale. Secondly, training sets contain only video-level labels indicating the existence of an abnormal event during the full video while lacking annotations of precise time durations. To tackle these problems, we contribute a new \textbf{L}arge-scale \textbf{A}nomaly \textbf{D}etection (\textbf{LAD}) database as the benchmark for anomaly detection in video sequences, which is featured in two aspects. 1) It contains 2000 video sequences including normal and abnormal video clips with 14 anomaly categories including~\emph{crash, fire, violence},~\emph{etc.} with large scene varieties, making it the largest anomaly analysis database to date. 2) It provides the annotation data, including video-level labels (abnormal/normal video, anomaly type) and frame-level labels (abnormal/normal video frame) to facilitate anomaly detection. Leveraging the above benefits from the LAD database, we further formulate anomaly detection as a fully-supervised learning problem and propose a multi-task deep neural network to solve it. We firstly obtain the local spatiotemporal contextual feature by using an Inflated 3D convolutional (I3D) network. Then we construct a recurrent convolutional neural network fed the local spatiotemporal contextual feature to extract the spatiotemporal contextual feature. With the global spatiotemporal contextual feature, the anomaly type and score can be computed simultaneously by a multi-task neural network. Experimental results show that the proposed method outperforms the state-of-the-art anomaly detection methods on our database and other public databases of anomaly detection. Supplementary materials are available at \url{http://sim.jxufe.cn/JDMKL/ymfang/anomaly-detection.html}.
	\end{abstract}
	
	\maketitle
	
	\section{Introduction}\label{sec1}
	
	Anomaly detection, which attempts to automatically predict abnormal/normal events in a given video sequence, has been actively studied in the field of computer vision. As a high-level computer vision task, anomaly detection aims to effectively distinguish abnormal and normal activities as well as anomaly categories in video sequences. In the last few years, there have been many studies investigating anomaly detection in the research community~\cite{Li2014Anomaly, Cosar2017Toward, Ravanbakhsh2017Abnormal, Hasan2016Learning, Liu2018GAN, wan2020weakly, park2020learning, zaheer2020old, Liu2019MarginLE}.
	
	Compared with normal behaviors, an event that rarely occurs or with low probability is generally considered as~\emph{anomaly}. In practice, it is difficult to build effective anomaly detection models due to the unknown event type and indistinct definition of~\emph{anomaly}. Traditionally, anomaly detection methods are designed from two aspects. One type of anomaly detection method is designed by reconstruction and they focus on modelling normal patterns in video sequences~\cite{Antic2011Video, Ravanbakhsh2017Abnormal, Hasan2016Learning, Lu2013Abnormal, Liu2018GAN, park2020learning, zaheer2020old}. The goal of these methods is to learn a feature representation model for normal patterns. At the testing stage, these methods utilize the differences between abnormal and normal samples to determine the final anomaly score of testing data, such as the reconstruction cost or specific threshold~\cite{Ravanbakhsh2017Abnormal, Hasan2016Learning, Lu2013Abnormal, Liu2018GAN, park2020learning, zaheer2020old}. Although reconstruction-based anomaly detection methods are good at reconstructing normal patterns in video sequences, the key issue in these methods is that they rely heavily on training data.
	
	\begin{figure*}[ht]
		\addtocounter{subfigure}{-6}
		\centering
		
		\subfigure{\label{fig:subfig:Case}
			\includegraphics[width = 2.6cm]{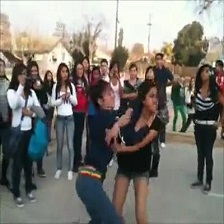}}
		\subfigure{\label{fig:subfig:Case}
			\includegraphics[width = 2.6cm]{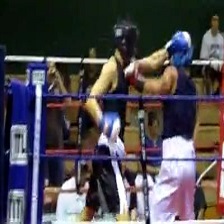}}
		\subfigure{\label{fig:subfig:Case}
			\includegraphics[width = 2.6cm]{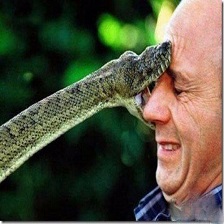}}
		\subfigure{\label{fig:subfig:Case}
			\includegraphics[width = 2.6cm]{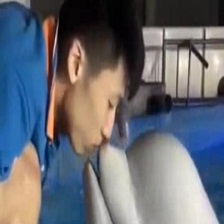}}
		\subfigure{\label{fig:subfig:Case}
			\includegraphics[width = 2.6cm]{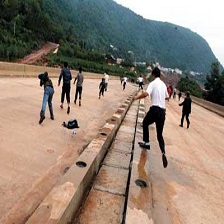}}
		\subfigure{\label{fig:subfig:Case}
			\includegraphics[width = 2.6cm]{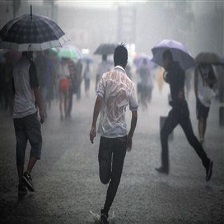}}
		
		\subfigure[\textit{Fighting}]{\label{fig:subfig:Case}
			\includegraphics[width = 2.6cm]{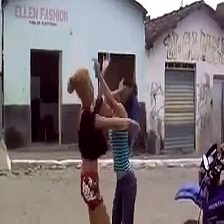}}
		\subfigure[\textit{Boxing}]{\label{fig:subfig:Case}
			\includegraphics[width = 2.6cm]{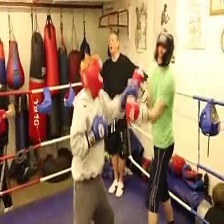}}
		\subfigure[\textit{Hurt}]{\label{fig:subfig:Case}
			\includegraphics[width = 2.6cm]{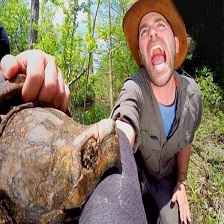}}
		\subfigure[\textit{Touching}]{\label{fig:subfig:Case}
			\includegraphics[width = 2.6cm]{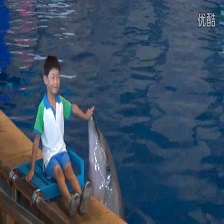}}
		\subfigure[\textit{Running}-A]{\label{fig:subfig:Case}
			\includegraphics[width = 2.6cm]{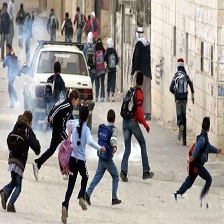}}
		\subfigure[\textit{Running}-N]{\label{fig:subfig:Case}
			\includegraphics[width = 2.6cm]{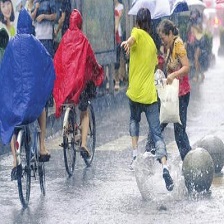}}
		
		\caption{Definition of anomaly on different visual scenes, where \textit{Running}-A and \textit{Running}-N are short for \textit{Running} Abnormal Events and \textit{Running} Normal Events, respectively. Column (a) to (f): \textit{Fighting}, \textit{Boxing}, \textit{Hurt}, \textit{Touching}, \textit{Running} Abnormal Events and \textit{Running} Normal Events.}
		\label{fig:Examples_of_case}
	\end{figure*}
	
	\begin{table*}[ht]
		\centering \caption{The detailed information of existing video anomaly detection databases.}
		\label{tab:basic_information_of_datasets}
		\renewcommand\tabcolsep{20pt} 
		\begin{tabular}{  c  c  c  c  c  c  }
			\toprule  
			Database&Year&Videos&Scenes&Supervision&Categories\\
			\midrule  
			UCSD Ped1~\cite{Li2014Anomaly} &2014&70&1&Video-level& not specified\\
			UCSD Ped2~\cite{Li2014Anomaly} &2014&28&1&Video-level& not specified\\
			Avenue~\cite{Lu2013Abnormal} &2013&37&1& Video-level& not specified\\
			LV~\cite{Leyva2017Video} &2017&28&28 & Video-level& not specified\\
			ShanghaiTech~\cite{Luo2017A} &2017&437&13& Video-level& not specified\\
			UCF-Crime~\cite{Sultani2018Real} &2018&1900&-& Video-level& 13\\
			\midrule  
			\textbf{LAD}&\textbf{-}&\textbf{2000}&\textbf{1895}& \textbf{Video- and frame- level}& \textbf{14}\\
			\bottomrule  
		\end{tabular}
		
	\end{table*}

	Another type of anomaly detection method regards anomaly detection as a classification problem~\cite{Zhu2013Anomaly, Colque2015Histograms}. For these methods, anomaly scores of video sequences are predicted by extracting features such as Histogram of Optical Flow (HOF) or dynamic texture (DT) with a trained classifier~\cite{Zhu2013Anomaly, Colque2015Histograms}. The performance of these methods is highly dependent on training samples. To obtain satisfactory performance, extracting effective and discriminative features is crucial for such anomaly detection methods.
	
	Most of the existing anomaly detection methods are designed based on the hypothesis that any pattern different from learned normal patterns is regarded as an anomaly. Under this assumption, the same activity in different scenes might be denoted as a normal or abnormal event. For example, as shown in Fig.~\ref{fig:Examples_of_case}, a fighting scene where two men are brawling may be considered as abnormal, while it may be normal when these two men are doing boxing sport; a girl/boy running on the street because of panic may be considered as abnormal, but it may be normal when the weather is raining since the girl/boy forget to take an umbrella; an animal touching human may be considered as abnormal (i.e., snake bite human), while it may be normal in the case of kissing people by a dolphin. Additionally, there is much redundant visual information in high-dimensional video data, which increases the difficulty for event representation in video sequences.
	
	The main challenges of anomaly detection task are caused by the lack of a large-scale anomaly detection database with fine-grained annotations. Although several anomaly detection databases are proposed in the research community~\cite{Li2014Anomaly, Lu2013Abnormal, Leyva2017Video, Sultani2018Real, Luo2017A}, they are flawed either in the dataset scale or annotation richness. Specifically, there are no more than 100 video sequences in ~\cite{Li2014Anomaly,Lu2013Abnormal,Leyva2017Video}, which could not satisfy the requirement of training data for deep learning based models. Besides, existing databases~\cite{Li2014Anomaly, Lu2013Abnormal, Leyva2017Video, Sultani2018Real, Luo2017A} only provide video-level labels in training set, which makes it infeasible to learn anomaly detection models in a fully-supervised manner. Moreover, the definition of anomaly is unclear, which makes it hard for anomaly ground-truth annotation and computational model design. For anomaly detection models for specific events such as hyperspectral anomaly detection~\cite{yuan2015hyperspectral}, violence detector~\cite{Mohammadi2016Angry} and traffic detector~\cite{Sultani2010Abnormal, yuan2016anomaly}, their applications are limited since they cannot be used to detect other abnormal events.
	
	To address these above problems in existing anomaly detection studies, we investigate anomaly detection from the following two aspects in this study.
	
	\begin{itemize}
		\item We build a new \textbf{L}arge-scale \textbf{A}nomaly \textbf{D}etection (\textbf{LAD}) database consisting of $2000$ video sequences and corresponding anomaly ground-truth data including video-level labels (abnormal/normal video, anomaly type) and frame-level labels (abnormal/normal video frame). There are 14 abnormal event categories in total. More than 100 video sequences are collected for each abnormal category, making it the largest database for anomaly detection to date.
		\item We propose a multi-task deep neural network for anomaly detection by learning local and global contextual spatiotemporal features with a multi-task joint learning scheme. An inflated 3D convolutional network is constructed to extract local spatiotemporal contextual features, which are further used to input a designed recurrent convolutional neural network to learn global spatiotemporal contextual features. The anomaly category and score can be predicted by a multi-task deep network with these global features.
	\end{itemize}
	
	The rest of this paper is organized as follows. Section II reviews the related work. Section III provides the details of the built large anomaly database, including the data collection and annotations. Section IV describes the proposed method in detail. Section V briefly describes the performance evaluation metrics and the performance comparison of the proposed method. We conclude this paper in Section VI.

	\section{Related Work}\label{sec_framework}
	
	\subsection{Anomaly Detection Databases}\label{Image-Based Saliency}
	
	Currently, there have been several anomaly detection databases for video sequences~\cite{Li2014Anomaly, Lu2013Abnormal,Leyva2017Video,Leyva2017Video, Sultani2018Real}. The detailed information of these existing databases is given in Table~\ref{tab:basic_information_of_datasets}.
	
	\textbf{UCSD}~\cite{Li2014Anomaly} includes two subsets of~\textbf{Ped1} and~\textbf{Ped2}, where an anomaly event is defined as a car or a bicycle appearing abnormally in the street compared with normal patterns of the car or pedestrian. In this database, the crowd density of different video sequences is different. All video sequences are with 10 Frames Per Second (fps), including two different outdoor scenes. The first subset~\textbf{Ped1} contains 34 training and 36 testing video sequences, including around 8000 video frames in total, while the second subset~\textbf{Ped2} contains 16 training and 12 testing video sequences including 4950 video frames in total. 
	
	\textbf{Avenue}~\cite{Lu2013Abnormal} contains 16 training and 21 testing video sequences. In this database, abnormal events are labeled as people running, loitering, throwing,~\emph{etc.} The size of a person may vary depending on the position and angle of the camera. It provides pixel-level annotation for each video frame. Each video sequence is about 2 minutes long. There are around 31000 video frames with a resolution of $640 \times 360$ in total. All video sequences are recorded in the same visual scene.
	
	\textbf{LV}~\cite{Leyva2017Video} This database contains 28 realistic video sequences for out-door and in-door scenes, and abnormal events are labeled as people fighting, people clashing, vandalism,~\emph{etc.} Each video sequence is divided into training and testing data. 
	
	\textbf{ShanghaiTech}~\cite{Luo2017A} contains 437 realistic video sequences for out-door scenes. There are 13 different visual scenes in this database, where all video sequences are captured by surveillance cameras. This database 130 abnormal events including Running, bicycles, skaters~\emph{etc.}
	
	\textbf{UCF-Crime}~\cite{Sultani2018Real} contains 13 real-world anomaly categories, including~\emph{Abuse, Arrest, Arson, Assault, Accident, Burglary, Explosion, Fighting, Robbery, Shooting, Stealing, Shoplifting} and~\emph{Vandalism}. It includes 1900 surveillance video sequences in total, composed of 950 abnormal video sequences and 950 normal video sequences. There are about 128 hours for all these video sequences in this database. The testing set including 150 normal and 140 abnormal video sequences, while the rest are used as the training set. This database provides only video-level labels for training videos.
	
	From Table~\ref{tab:basic_information_of_datasets}, we can observe that the training video of most existing anomaly detection databases is limited in scale. Although they contain a variety of abnormal events, the categories of abnormal videos are not specified. However, the visual scenes in the real world are diverse and complicated with different anomaly types. Another common drawback of existing databases is the lacking of frame-level labels. As a result, anomaly detection algorithms can only be learned in a weakly-supervised manner, which deteriorates the performance and impedes the wide usage in practical applications.
	
	In this work, we build a new large-scale anomaly detection database, including 2000 video sequences and the corresponding video- and frame-level labels, to promote anomaly detection in a fully-supervised manner. The built database contains 1895 different visual scenes with 14 anomaly categories, including~\emph{Crash, Crowd, Destroy, Drop, Falling, Fighting, Fire, Fall Into Water, Hurt, Loitering, Panic, Thiefing, Trampled,} and~\emph{Violence}. We will introduce this database in detail in Section III.
	
	\subsection{Anomaly Detection Methods}\label{Video-Based Saliency}
	
	Early anomaly detection studies extract object trajectories to detect abnormal activities in video sequences, where an object against the learned normal object trajectories is detected as an anomaly~\cite{Cosar2017Toward, Piciarelli2008Trajectory, Wu2010Chaotic, piciarelli2006on-line, jiang2011anomalous, tung2011goal-based, morris2011trajectory, calderara2011detecting, patino2015abnormal, yi2015understanding}. Cosar~\emph{et al.} proposed an unsupervised architecture for abnormal behavior prediction by object trajectory analysis (i.e., speed, direction, and body movement) and pixel-level analysis (i.e., appearance)~\cite{Cosar2017Toward}. Piciarelli~\emph{et al.} designed an anomaly detection model by clustering extracted normal trajectories of moving objects in video sequences~\cite{Piciarelli2008Trajectory, piciarelli2006on-line}. Specifically, they utilized a single-class SVM to learn normal object trajectories. In the testing stage, a new trajectory is predicted as an anomaly or not by comparing it with the clustering model with a threshold. Wu~\emph{et al.} exploited chaotic invariants of lagrangian particle trajectories to represent anomaly activities in crowded scenes~\cite{Wu2010Chaotic}. Patino~\emph{et al.} detected speed and direction change by trajectories of moving objects, which predict anomaly events~\cite{patino2015abnormal}. Jiang~\emph{et al.} proposed a context-aware anomaly detection method~\cite{jiang2011anomalous}. By tracking all moving objects in a video sequence, the anomaly event is detected by considering different levels of spatiotemporal contexts. Morris~\emph{et al.} studied the features of the normal recurrent motion patterns of the surveillance subjects to detect abnormalities~\cite{morris2011trajectory}. Yi~\emph{et al.} proposed a pedestrian behavior model for anomaly detection by stationary crowd group~\cite{yi2015understanding}. However, these methods can not work well when objects are occlusive.
	
	To solve the challenging problem from object occlusion, some studies used global features to represent complex scenes for anomaly detection~\cite{Li2014Anomaly, Antic2011Video, Leyva2017Video, Wang2014Detection, Mehran2009Abnormal, adam2008robust, saligrama2012video, benezeth2009abnormal, kim2009observe, kratz2009anomaly, zhang2005semi-supervised, roshtkhari2013online, zhu2013context-aware, xiao2015learning, Cui2011Abnormal, Yuan2015Online, Cheng2015Gaussian}. Then they used a nonlinear one-class support vector machine to learn normal patterns. The event behavior with an outlier score predicted by the trained model is considered an an anomaly. Different from the study~\cite{Wang2014Detection}, Li~\emph{et al.} proposed a joint anomaly detection model by combining temporal and spatial anomalies with a Mixture of Dynamic Textures (MDT) for modelling normal crowd activities~\cite{Li2014Anomaly}. Besides, Mehran~\emph{et al.} introduced a social force model to stimulate the normal behaviour of the crowd. Then they classified video frames as normal or abnormal by using a bag of words approach~\cite{Mehran2009Abnormal}. Cui~\emph{et al.} defined a concept of interaction energy to represent the current interaction between the surrounding region and objects. A behaviour is considered as anomaly when the energy and velocity of an object change dramatically~\cite{Cui2011Abnormal}. Adam~\emph{et al.} used low-level information based on multiple local monitors for anomaly detection in video sequences~\cite{adam2008robust}. In order to detect abnormal events in video sequences, Saligrama~\emph{et al.} used spatiotemporal features with a $k$-nearest neighbor method to design an anomaly detection model~\cite{saligrama2012video}. Benezeth~\emph{et al.} used normal events to train a spatiotemporal co-occurrence matrix and used the matrix and Markov random field to detect anomaly~\cite{benezeth2009abnormal}. Kim~\emph{et al.} used a mixture of probabilistic PCA models to present the local optical flow pattern, and used the representation and Markov random field to define normal patterns~\cite{kim2009observe}. Antic~\emph{et al.} introduced a probabilistic model by localizing abnormalities with statistical inference~\cite{Antic2011Video}. Yuan~\emph{et al.} proposed an informative Structural Context Descriptor (SCD) to describe the crowd scene for anomaly detection~\cite{Yuan2015Online}. Lu~\emph{et al.} proposed to learn multiple dictionaries to model normal patterns with sparse constraint~\cite{Lu2013Abnormal}. Leyva~\emph{et al.} designed an anomaly detection method based on optical flow information and foreground occupancy~\cite{Leyva2017Video}. In~\cite{athanesious2020detecting}, a novel hand-craft optical-optical feature extractor named Super Orientation Optical Flow (SOOF) is proposed to efficiently capture motion information of objects in surveillance videos. In~\cite{yuan2015hyperspectral}, a vertex- and edge-weighted graph is constructed to reduce false-positive rate in hyperspectral anomaly detection task. To tackle specific problems caused by dynamic outdoor environments in traffic scenes, Yuan~\emph{et al.} proposed a spatial localization constrained sparse coding approach as a motion descriptor.

	Recently, deep learning techniques have been widely used to build anomaly detection models~\cite{Ravanbakhsh2017Abnormal, Hasan2016Learning, Liu2018GAN, Sultani2018Real, Sabokrou2017Deep, sabokrou2016video, Luo2017Remembering, Hinami2017Joint, Luo2017A, Ionescu2017Unmasking, Xu2015learning}. Sabokrou~\emph{et al.} proposed a cascaded Deep Neural Networks (DNN) for anomaly detection by hierarchically modelling normal patches using deep features, then they used Gaussian classifier to identify abnormal behaviours in video sequences~\cite{Sabokrou2017Deep}. Ravanbakhsh~\emph{et al.} trained two Generative Adversarial Nets (GANs) to learn normal patterns in video sequences~\cite{Ravanbakhsh2017Abnormal}. During the training stage, the first generator of GANs takes a normal video frame as input and produces a reconstructed optical flow image, while the second generator of GANs is fed into a real optical-flow image and generates a reconstructed appearance image. In the testing stage, this model detects anomalies by using the reconstruction differences between real data (original video frames and original optical-flow images) and generated data (reconstructed video frames and reconstructed optical-flow images. Hasan~\emph{et al.} proposed two auto-encoder models to learn temporal regularity for anomaly detection~\cite{Hasan2016Learning}. Similarly, Xu~\emph{et al.} proposed a deep neural network based model by constructing a stacked denoising autoencoder for feature learning for abnormal event detection~\cite{Xu2015learning}. Luo~\emph{et al.} proposed a Convolutional LSTMs Auto-Encoder (ConvLSTM-AE) to encode normal appearance and motion patterns for abnormal event detection~\cite{Luo2017Remembering}. Hinami~\emph{et al.} learned a Convolutional neural network through multiple visual tasks, then they used semantic information to detect anomaly events~\cite{Hinami2017Joint}. Ionescu~\emph{et al.} applied the unmasking technique to train a binary classifier to distinguish two consecutive short video sequences and gradually remove the most discriminant features~\cite{Ionescu2017Unmasking}. Luo~\emph{et al.} proposed a Temporally-coherent Sparse Coding (TSC) approach for anomaly detection, in which similar adjacent frames are encoded with similar reconstruction coefficients~\cite{Luo2017A}. Liu~\emph{et al.} proposed an anomaly detection model based on the difference between a predicted frame and the ground-truth, where the temporal constraint is considered besides spatial constraints~\cite{Liu2018GAN}. Sultani~\emph{et al.} learned a generic model using deep Multiple Instance Learning (MIL) framework with weakly labeled data~\cite{Sultani2018Real}, and Wan~\emph{et al.} proposed a dynamic MIL loss and a center loss for enlarging the inter-class distance between anomalous and normal instances and reducing the intra-class distance of normal instances, respectively.
	
	All above deep learning based methods formulate anomaly detection as the unsupervised learning or weakly-supervised learning problem due to the lack of frame-level labels in the training set of the existing anomaly detection databases. In this paper, leveraging the fine-grained frame-level annotation from our proposed LAD database, we formulate anomaly detection as a fully-supervised learning problem and propose a novel multi-task deep neural network to address anomaly detection in videos. Through extensive experimental analysis, we show that our model significantly improves the performance anomaly detection.

	\section{Anomaly Detection Benchmark}
	
	\subsection{Data Collection}
	\begin{figure*}[htb]
		
		\centering
		\begin{minipage}{1\linewidth}
			\centerline{
				\tiny \rotatebox{90}{\qquad\qquad \textbf{Crash}}
				\hspace{0.1cm}
				\includegraphics[width=0.115\textwidth]{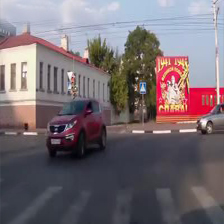}
				\includegraphics[width=0.115\textwidth]{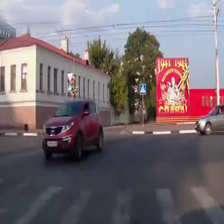}
				\includegraphics[width=0.115\textwidth]{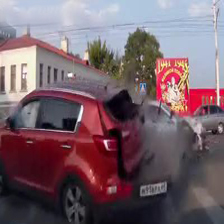}
				\includegraphics[width=0.115\textwidth]{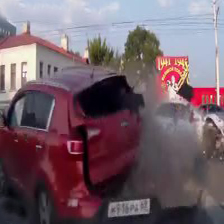}
				\vspace{0.1cm}
				\hspace{0.1cm}
				\tiny \rotatebox{90}{\qquad\qquad \textbf{Crowd}}
				\hspace{0.1cm}
				\includegraphics[width=0.115\textwidth]{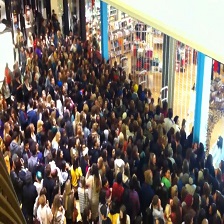}
				\includegraphics[width=0.115\textwidth]{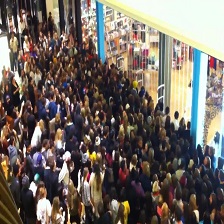}
				\includegraphics[width=0.115\textwidth]{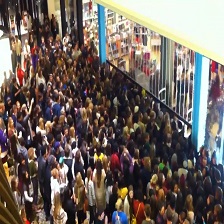}
				\includegraphics[width=0.115\textwidth]{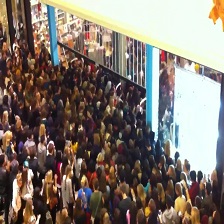}
				\vspace{0.1cm}
			}
		\end{minipage}

		\begin{minipage}{1\linewidth}
			\centerline{
				\tiny \rotatebox{90}{\qquad\qquad \textbf{Destroy}}
				\hspace{0.1cm}
				\includegraphics[width=0.115\textwidth]{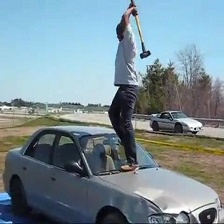}
				\includegraphics[width=0.115\textwidth]{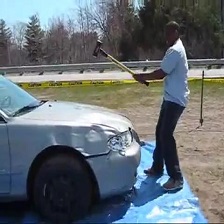}
				\includegraphics[width=0.115\textwidth]{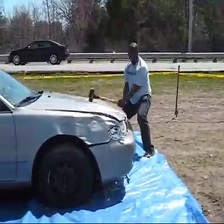}
				\includegraphics[width=0.115\textwidth]{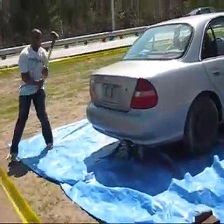}
				\vspace{0.1cm}
				\hspace{0.1cm}
				\tiny \rotatebox{90}{\qquad\qquad \textbf{Drop}}
				\hspace{0.1cm}
				\includegraphics[width=0.115\textwidth]{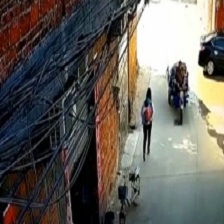}
				\includegraphics[width=0.115\textwidth]{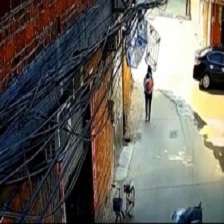}
				\includegraphics[width=0.115\textwidth]{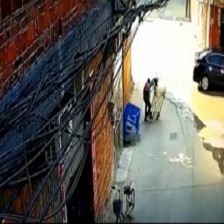}
				\includegraphics[width=0.115\textwidth]{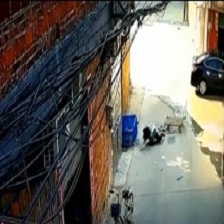}
				\vspace{0.1cm}
			}
		\end{minipage}
		
		\begin{minipage}{1\linewidth}
			\centerline{
				\tiny \rotatebox{90}{\qquad \textbf{Falling}}
				\hspace{0.1cm}
				\includegraphics[width=0.115\textwidth]{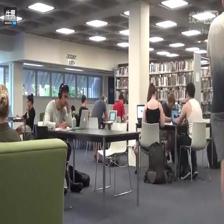}
				\includegraphics[width=0.115\textwidth]{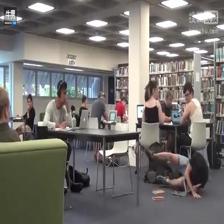}
				\includegraphics[width=0.115\textwidth]{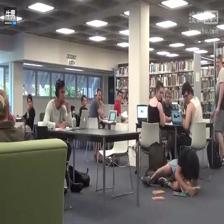}
				\includegraphics[width=0.115\textwidth]{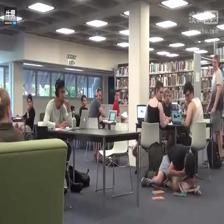}
				\vspace{0.1cm}
				\hspace{0.1cm}
				\tiny \rotatebox{90}{\qquad \textbf{FallIntoWater}}
				\hspace{0.1cm}
				\includegraphics[width=0.115\textwidth]{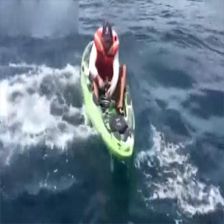}
				\includegraphics[width=0.115\textwidth]{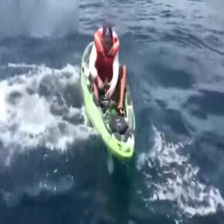}
				\includegraphics[width=0.115\textwidth]{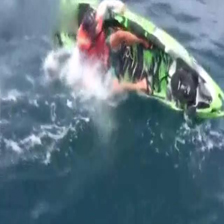}
				\includegraphics[width=0.115\textwidth]{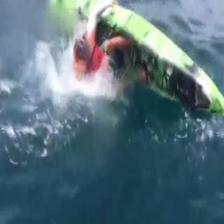}
				\vspace{0.1cm}
			}
		\end{minipage}
		
		\begin{minipage}{1\linewidth}
			\centerline{
				\tiny \rotatebox{90}{\qquad \textbf{Fighting}}
				\hspace{0.1cm}
				\includegraphics[width=0.115\textwidth]{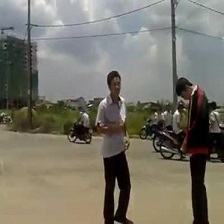}
				\includegraphics[width=0.115\textwidth]{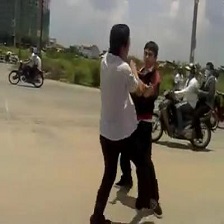}
				\includegraphics[width=0.115\textwidth]{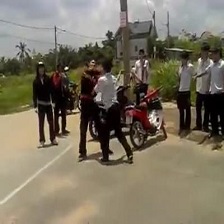}
				\includegraphics[width=0.115\textwidth]{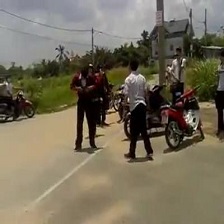}
				\vspace{0.1cm}
				\hspace{0.1cm}
				\tiny \rotatebox{90}{\qquad\qquad \textbf{Fire}}
				\hspace{0.1cm}
				\includegraphics[width=0.115\textwidth]{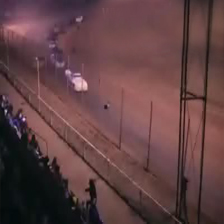}
				\includegraphics[width=0.115\textwidth]{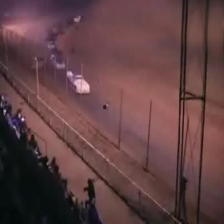}
				\includegraphics[width=0.115\textwidth]{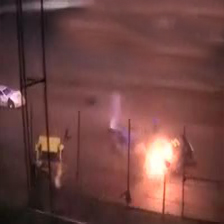}
				\includegraphics[width=0.115\textwidth]{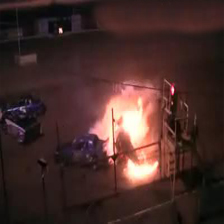}
				\vspace{0.1cm}
			}
		\end{minipage}
		
		\begin{minipage}{1\linewidth}
			\centerline{
				\tiny \rotatebox{90}{\qquad\qquad \textbf{Hurt}}
				\hspace{0.1cm}
				\includegraphics[width=0.115\textwidth]{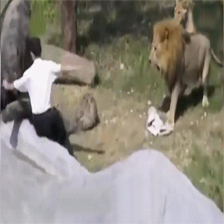}
				\includegraphics[width=0.115\textwidth]{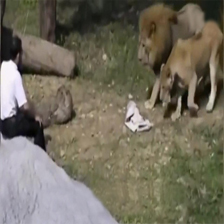}
				\includegraphics[width=0.115\textwidth]{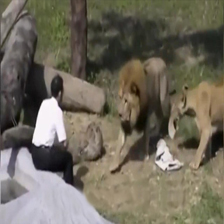}
				\includegraphics[width=0.115\textwidth]{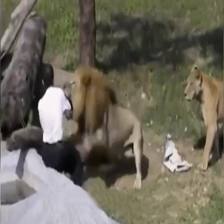}
				\vspace{0.1cm}
				\hspace{0.1cm}
				\tiny \rotatebox{90}{\qquad\qquad \textbf{Loitering}}
				\hspace{0.1cm}
				\includegraphics[width=0.115\textwidth]{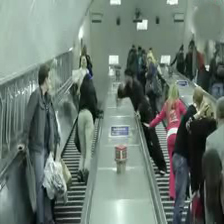}
				\includegraphics[width=0.115\textwidth]{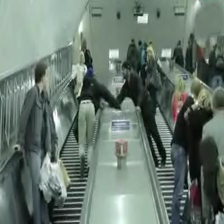}
				\includegraphics[width=0.115\textwidth]{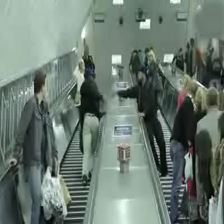}
				\includegraphics[width=0.115\textwidth]{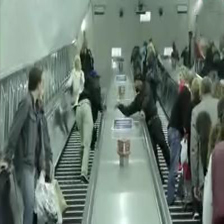}
				\vspace{0.1cm}
			}
		\end{minipage}
		
		\begin{minipage}{1\linewidth}
			\centerline{
				\tiny \rotatebox{90}{\qquad\qquad \textbf{Panic}}
				\hspace{0.1cm}
				\includegraphics[width=0.115\textwidth]{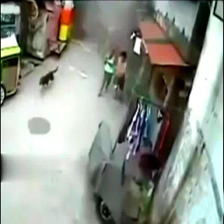}
				\includegraphics[width=0.115\textwidth]{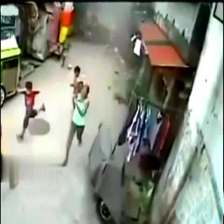}
				\includegraphics[width=0.115\textwidth]{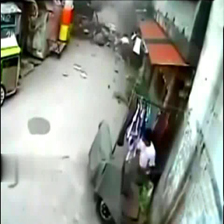}
				\includegraphics[width=0.115\textwidth]{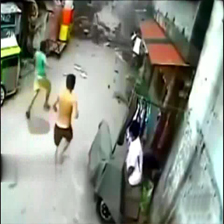}
				\vspace{0.1cm}
				\hspace{0.1cm}
				\tiny \rotatebox{90}{\qquad\qquad \textbf{Thiefing}}
				\hspace{0.1cm}
				\includegraphics[width=0.115\textwidth]{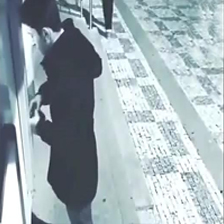}
				\includegraphics[width=0.115\textwidth]{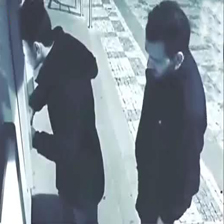}
				\includegraphics[width=0.115\textwidth]{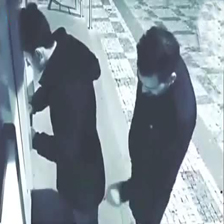}
				\includegraphics[width=0.115\textwidth]{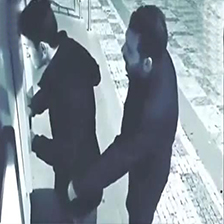}
				\vspace{0.1cm}
			}
		\end{minipage}
		
		\begin{minipage}{1\linewidth}
			\centerline{
				\tiny \rotatebox{90}{\qquad\qquad \textbf{Trampled}}
				\hspace{0.1cm}
				\includegraphics[width=0.115\textwidth]{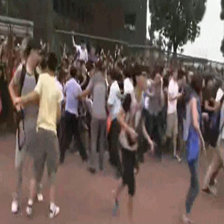}
				\includegraphics[width=0.115\textwidth]{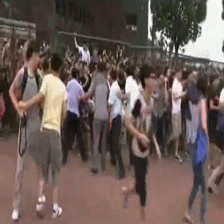}
				\includegraphics[width=0.115\textwidth]{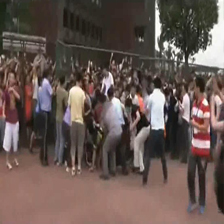}
				\includegraphics[width=0.115\textwidth]{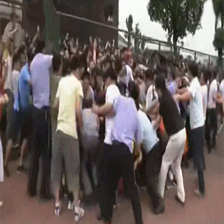}
				\vspace{0.1cm}
				\hspace{0.1cm}
				\tiny \rotatebox{90}{\qquad\qquad \textbf{Violence}}
				\hspace{0.1cm}
				\includegraphics[width=0.115\textwidth]{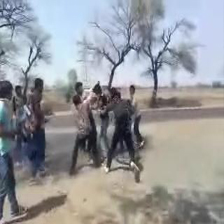}
				\includegraphics[width=0.115\textwidth]{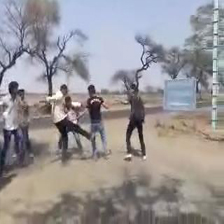}
				\includegraphics[width=0.115\textwidth]{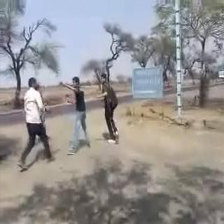}
				\includegraphics[width=0.115\textwidth]{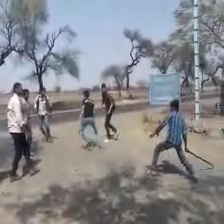}
				\vspace{0.1cm}
			}
		\end{minipage}
		
		\begin{minipage}{1\linewidth}
			\centerline{
				\tiny \rotatebox{90}{\qquad \textbf{Normal (I)}}
				\hspace{0.1cm}
				\includegraphics[width=0.115\textwidth]{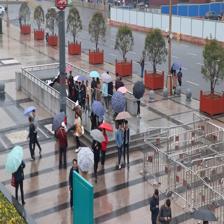}
				\includegraphics[width=0.115\textwidth]{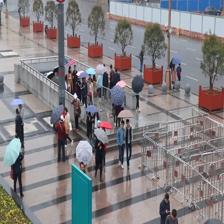}
				\includegraphics[width=0.115\textwidth]{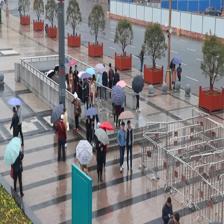}
				\includegraphics[width=0.115\textwidth]{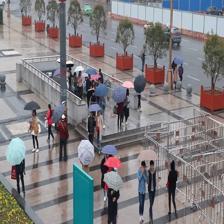}
				\vspace{0.1cm}
				\hspace{0.1cm}
				\tiny \rotatebox{90}{\qquad \textbf{Normal (II)}}
				\hspace{0.1cm}
				\includegraphics[width=0.115\textwidth]{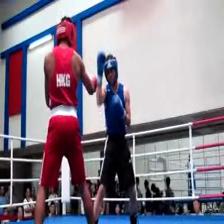}
				\includegraphics[width=0.115\textwidth]{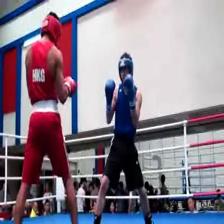}
				\includegraphics[width=0.115\textwidth]{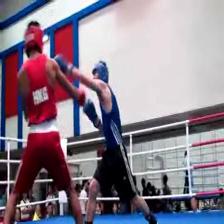}
				\includegraphics[width=0.115\textwidth]{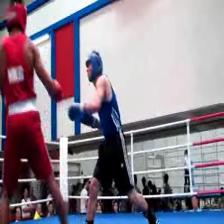}
				\vspace{0.1cm}
			}
		\end{minipage}
		
		\caption{Visual samples of different anomaly event categories in the proposed LAD database. The proposed database contains 14 distinct anomaly categories, including~\emph{Crash, Crowd, Destroy, Drop, Falling, Fall Into Water, Fighting, Fire, Hurt, Loitering, Panic, Thiefing, Trampled and Violence,} as well as~\emph{Normal} activities.}
		
		\label{fig:Examples_of_dataset}
	\end{figure*}
	
	To collect large-scale representative anomaly activities, we search for a large number of video sequences from public websites including YouTube\footnote{https://www.youtube.com/}, YouKu\footnote{https://www.youku.com/}, and Tencent Video\footnote{https://v.qq.com/}. Besides, we collect some video sequences from existing activity recognition databases, such as FCVID~\cite{FCVID}, Hollywood2~\cite{Marszalek2009Actions}, and YouTube Action~\cite{Liu2009Recognizing}. Additionally, we record some normal activities or suddenly occurring abnormal events in the square and school by a digital camera to provide plenty of visual scenes and real-world events. With these operations, we initially collect over 2500 video sequences in total.
	
	We analyze the collected video sequences and classify these video sequences into 14 categories, including~\emph{Crash, Crowd, Destroy, Drop, Falling, Fighting, Fire, Fall Into Water, Hurt, Loitering, Panic, Thiefing, Trampled,} and~\emph{Violence}. For each category, we discard some video sequences which fall into any of the following two conditions: (1) low resolution or low quality; and (2) incomplete anomaly event or anomaly is not clear. We strictly select more than 100 video sequences, including more than 50 normal video sequences and 50 abnormal video sequences for each category. Finally, we preserve 14 distinct anomaly categories with 2000 video sequences totally. The frame rate of all video sequences is 25 fps. For each video sequence, we manually extract a video segment that represents an abnormal/normal activity by irrelevant video frames. In Fig.~\ref{fig:Examples_of_dataset}, we show four frames of an example video for each anomaly category, including 2 normal frames and 2 abnormal frames.
	
	\subsection{Annotations}

	\begin{figure*}[ht]
		\centering
		\subfigure[]{\label{fig:dataset_info}
			\includegraphics[width = 5.4cm]{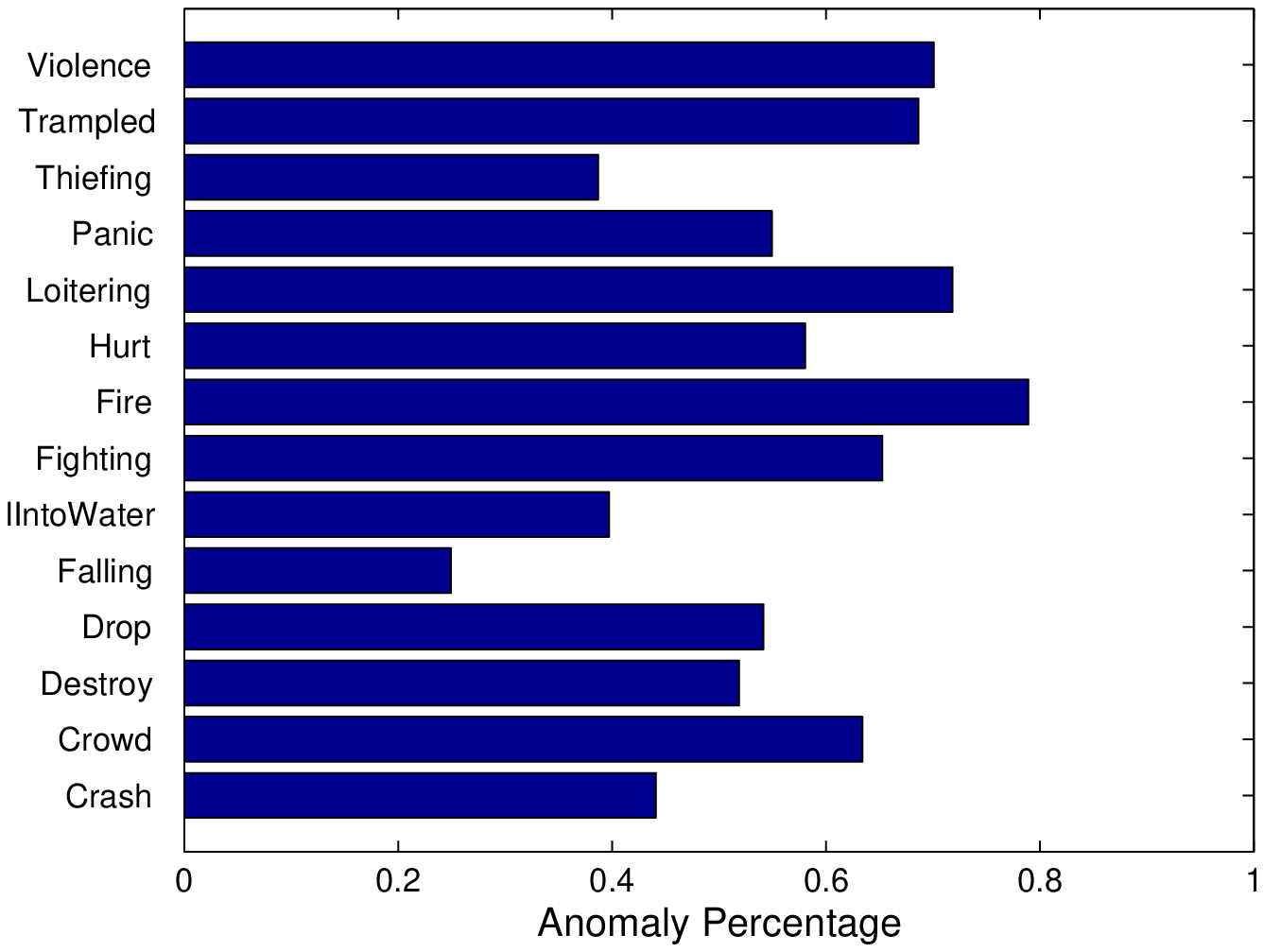}}
		\subfigure[]{\label{fig:dataset_info}
			\includegraphics[width = 5.4cm]{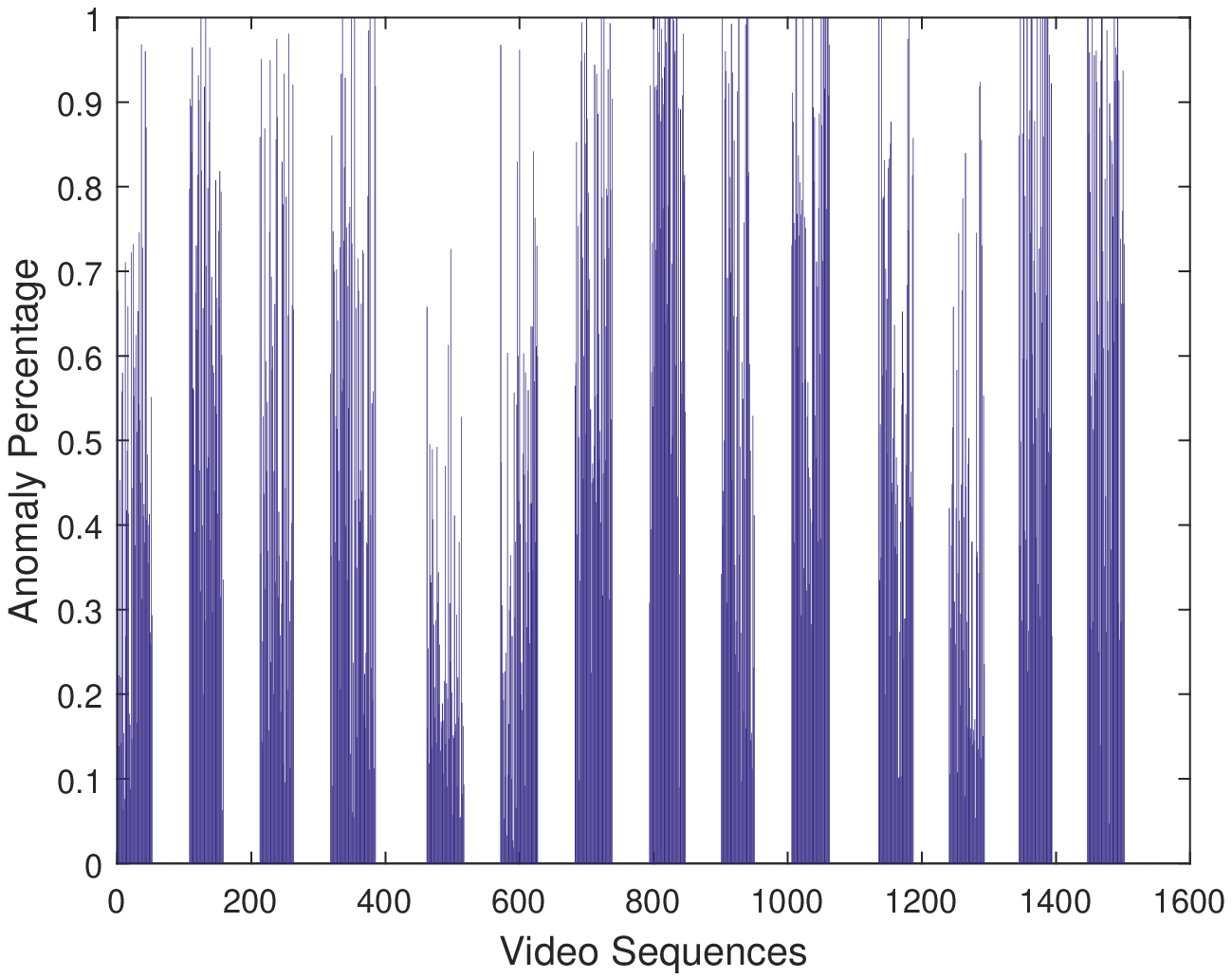}}
		\subfigure[]{\label{fig:dataset_info}
			\includegraphics[width = 5.4cm]{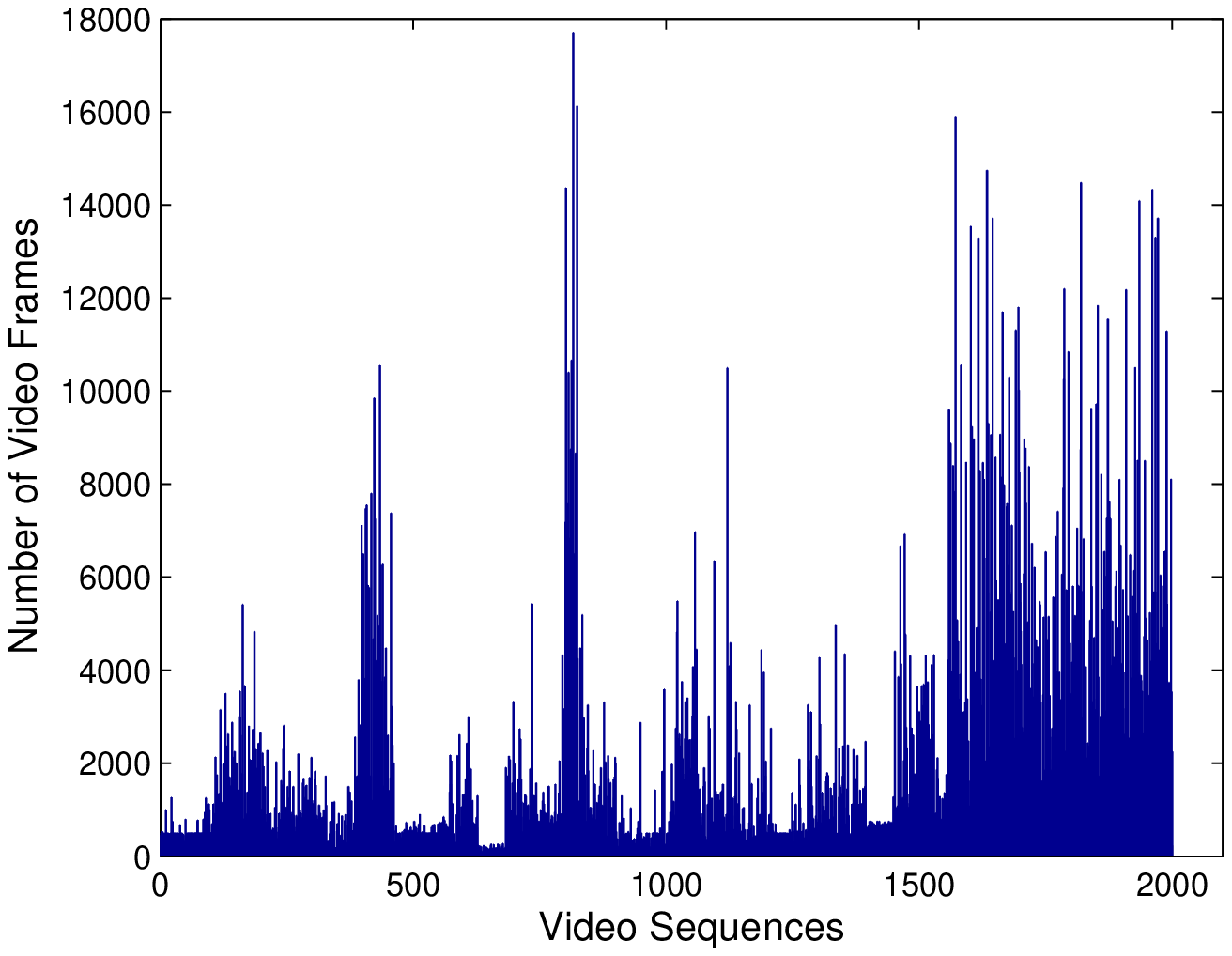}}

		\caption{The statistics information of the proposed LAD database. (a) The anomaly distribution of each anomaly category; (b) the anomaly distribution of each video sequence; (c) the number of video frames for each video sequence.}
		\label{fig:dataset_info}
	\end{figure*}

	As a high-level video analysis task, anomaly detection requires frame-level labels to identify the time period of an abnormal event starts and video-level labels to recognize the anomaly category. Thus, we provide both video- and frame-level labels in our database. To ensure the quality of annotations, we invite five postgraduate students to take part in our annotation experiment. In the annotation experiment, we define 1 as abnormal video frame and 0 as normal video frame. We first ask annotators to find the video frames where an anomaly event begins and ends, which are all labeled as 1, and the rest are labeled as 0. Then, we compute the averaging scores of annotations for each frame. Finally, we binaries the averaging scores by using threshold 0.5, and take binary averaging scores as the frame-level anomaly labels. Video-level labels are annotated to represent anomaly category, where a video sequence is labeled anomaly if any frame in this video sequence is abnormal.
	
	In this database, a normal video sequence in each anomaly category denotes that behaviour in this video sequence is regarded as normal. For example, for the~\emph{\textit{Fighting}} category, the boxing activity is classified as normal though it is similar to the fighting anomaly event; for the~\emph{Falling} category, a woman falling down when playing roller-skating is labeled as anomaly, while a woman bending into a squat with knees is annotated as normal; for the~\emph{Hurt} category, a woman being attacked by a dog is labeled as anomaly, while a woman walking the dog is annotated as normal. We divide the built database into training and testing subsets. The testing set contains 560 sequences, composed of randomly selected 20 abnormal and 20 normal video sequences for each anomaly category. The rest are used as training set. The statistics information of all video sequences is shown in Fig.~\ref{fig:dataset_info}. In the built database, we record the entire process from the beginning to the end for anomaly events, and we use a video sequence to represent a complete event. As shown in Fig.~\ref{fig:dataset_info}, the number of video frames for most video sequences is in the range of $\left[ 4000, 8000 \right]$. The anomaly percentage of \textit{Fire} and \textit{Loitering} event categories are high because the anomaly of these anomaly events generally lasts for a long time. The video frames with smoke or small fires are considered anomalies when we annotate abnormal fire frames for \textit{Fire}. By contrast, the anomaly percentage of the \textit{Falling} category is the lowest since this type of anomaly event lasts for a short time. In this type of video sequences, when a person falls down, he can stand up quickly. Besides, we compare our database with UCF-Crime~\cite{Sultani2018Real}, and find that there are some video sequences with an anomaly percentage higher than 0.5. Since the anomaly events of these video sequences last for a long time, the whole event can be fully expressed. In addition, the abnormal frames of UCF-Crime database~\cite{Sultani2018Real} are not completely labeled. For example, the authors only consider the moment of explosion as abnormal for the Explosion category, but the fire generated after the explosion is regarded as normal.
	
	\section{Proposed Method}\label{sec_framework}
	\begin{figure*}[ht]
		\centering
		\includegraphics[width = 0.96\textwidth]{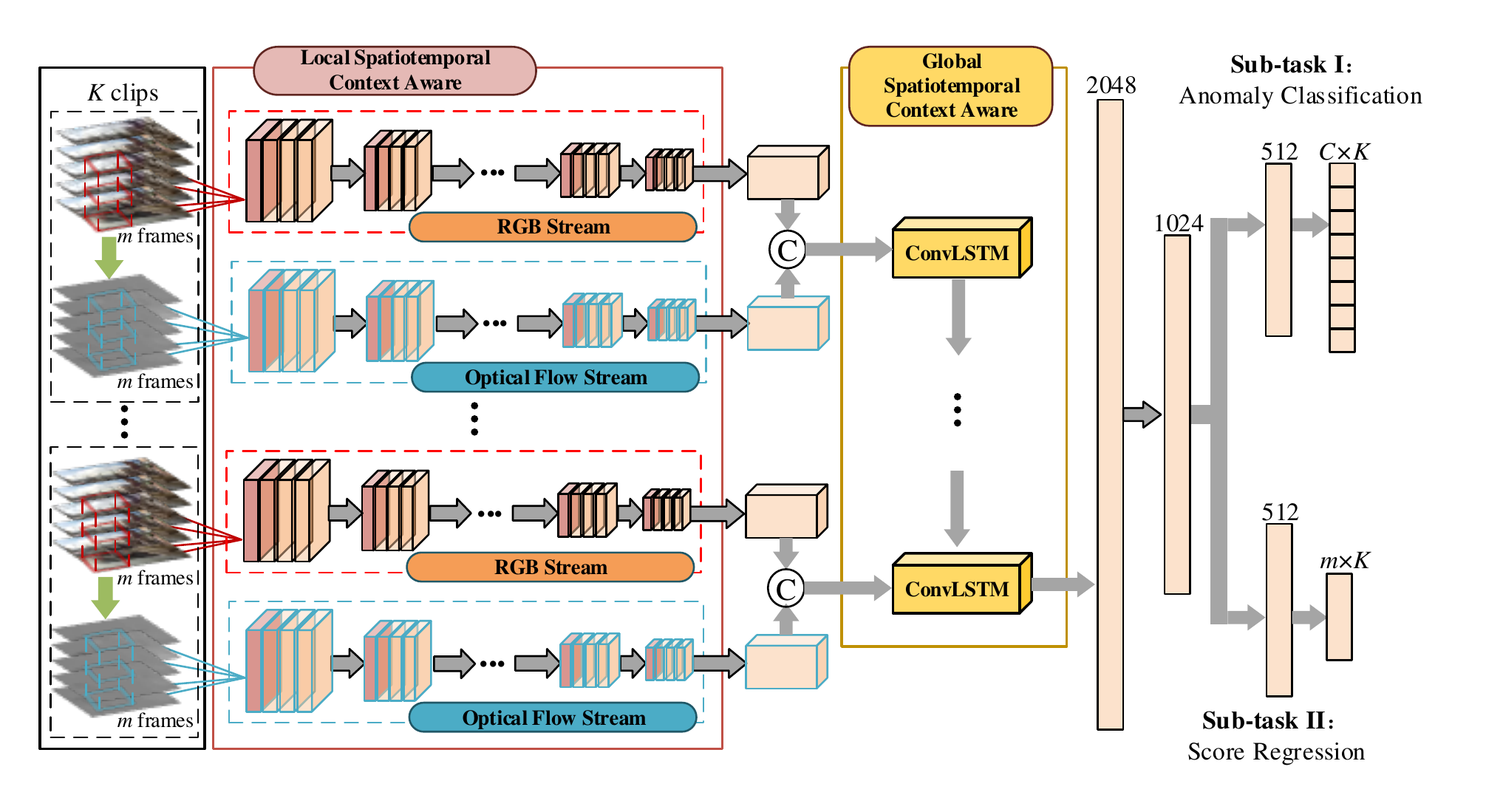}
		\caption{The architecture of the proposed anomaly detection method by modeling local and global spatiotemporal contextual features.}
		\label{fig:framework}
	\end{figure*}
	
	Here, we propose a multi-task deep neural network for anomaly detection. The proposed model is demonstrated in Fig.~\ref{fig:framework}. It consists of two components, i.e., a local and a global spatiotemporal context-aware streams.
	
	Our observation is that local outliers may failed to extract feature representation of continuous action. To alleviate this problem, we devise a local spatiotemporal context aware submodule and a spatiotemporal context aware submodule, as shown in Fig.~\ref{fig:framework}. In particular, we first encode each video sequence by feature representation with a pretrained Inflated 3D convolutional network (I3D)~\cite{CarreiraZ17}. Given a video sequence with $M$ frames, we divide it into $N$ clips, and each clip contains $m$ video frames. Thus, each video sequence can be denoted as ${\bm V} = \left\{{{\bm v}_{n}, N=M/m}\right\}_{n=1}^N$. The split clips are fed into a pretrained I3D to extract high-level visual features. For $K$ consecutive clips, the local feature vectors can be represented as $\bm{X}= \left\{{\bm x}_{t}\right\}_{t=1}^K$.
	
	The video sequence is high dimensional data that contains plenty of visual information. Thus, preserving important cues yet filtering out the redundancy is important to learn effective anomaly detection models. To learn robust global spatiotemporal contextual cues, we feed the extracted local contextual features of $K$ consecutive clips into the global context-aware stream to learn high-level features. As shown in Fig.~\ref{fig:framework}, we adopt a two-layer Convolutional LSTM (ConvLSTM) network~\cite{Shi2015Convolutional} to learn global spatiotemporal features of a video segment. Unlike LSTM, ConvLSTM~\cite{Shi2015Convolutional} is designed by using three-dimensional data as the input and uses convolutional operation, which can obtain temporal information and extract spatial features. At the same time, it provides good generalization by reducing the number of parameters and the computational complexity. Specifically, we show the formula for ConvLSTM as follows.
	
	\begin{equation}
	{\bm{{i}}}_{t}=\sigma (\bm{W}_{xi} * \bm{X}_{t}+ \bm{W}_{hi} * \bm{H}_{t-1} + \bm{W}_{ci} \circ \bm{C}_{t-1}+b_{i})
	\end{equation}
	\begin{equation}
	{\bm{{f}}}_{t}=\sigma (\bm{W}_{xf} * \bm{X}_{t} + \bm{W}_{hf} * \bm{H}_{t-1} + \bm{X}_{cf} \circ \bm{C}_{t-1}+b_{f})
	\end{equation}
	\begin{equation}
	\bm{C}_{t}={\mathclap{\bm{{f}}}_{t}}{\circ}{\bm{C}_{t-1}}+{\mathclap{\bm{{i}}}_{t}} \circ {\rm tanh}(\bm{W}_{xc} * \bm{X}_{t}+\bm{X}_{hc} * \bm{H}_{t-1}+b_{c})
	\end{equation}
	\begin{equation}
	{\bm{{o}}}_{t}=\sigma (\bm{W}_{xo} * \bm{X}_{t} + \bm{W}_{ho} * \bm{H}_{t-1} + \bm{W}_{co} \circ \bm{C}_{t}+b_{o})
	\end{equation}
	\begin{equation}
	\bm{H}_{t}={\bm{{o}}}_{t} \circ {\rm tanh}(\bm{C}_{t})
	\end{equation}
	where ${\bm{{X}}}_{t}$ and $\bm{H}_{t}$ denote the input and output of ConvLSTM~\cite{Shi2015Convolutional} at time $t$; ${\bm{{i}}}_{t}$, ${\bm{{f}}}_{t}$, ${\bm{{o}}}_{t}$ and $\bm{C}_{t}$ represent outputs of input gate, forget gate, output gate and memory cell; $*$ is a convolutional operation; $\circ$ represents the Hadamard product; and $\sigma$ is the sigmoid activation function.
	
	For ConvLSTM, we use the local features extracted from each clip of a video segment as the input. The ConvLSTM network leverages both long- and short-term cues of input features. The hidden states of the last layer of ConvLSTM are fed into three fully convolutional layers to predict the final event category and anomaly scores.
	
	Furthermore, we design a multi-task joint learning network for learning the intrinsic relationship between anomaly detection and classification. The sub-network of anomalous categories classification task is designed to recognize anomaly category, and we use a cross-entropy loss function in this sub-network.
	
	\begin{equation}
	\emph{$L$}_{1}=-{\sum_{i=1}^C}{\hat{y}_{i}}\log{{y}_{i}}  + \gamma\parallel \bm{W} \parallel_{2}^{2}
	\end{equation}
	where ${{\hat{\bm y}}}=[\hat{y}_{1},\hat{y}_{2},...,\hat{y}_{C}]$ denotes the one hot encoding of anomalous category label for a video sequence; ${\bm y}=[{y}_{1},{y}_{2},...,{y}_{C}]$ represents the corresponding score vector predicted by the sub-network of anomalous categories classification; $\parallel {\bm W} \parallel_{2}^{2}$ is a $L_{2}$-norm regularization term to avoid over-fitting; $\gamma$ is a hyper-parameter to balance the trade-off between the loss and regularization.
	
	As the sub-task of anomaly score prediction is modeled as a regression problem. We use \textbf{$smooth$} loss function~\cite{Girshick2015Fast} as learning objective in this sub-network as follows.
	
	\begin{equation}\label{eq5}
	\begin{split}
	&~\emph{$L$}_{2} = \sum_{i} ({smooth}(s_{i}-\hat{s_{i}})) \\
	& 	\emph{smooth}(x) =
	\begin{cases}0.5x^{2}, &\left |x \right |\leq 1 \cr \left |x \right|-0.5, &otherwise
	\end{cases}
	\end{split}
	\end{equation}
	where $\hat{s_{i}}$ denotes the anomalous label of a video frame; ${s_{i}}$ represents the corresponding score predicted by the sub-network of anomaly score prediction. Based on ${L}_{1}$ and ${L}_{2}$, the final loss function is written as follows:

	\begin{equation}
	\emph{$L$} = {\lambda}_{1}{L}_{1} + {\lambda}_{2}{L}_{2}
	\end{equation}
	where $\lambda_{1}$ and $\lambda_{2}$ are hyper-parameters to weight the importance of two sub-task.
	
	\section{Experimental Results}\label{sec_experiment}

	\begin{table*}[ht]
		\centering 
		\renewcommand\tabcolsep{9pt} 
		\caption{Data splits on Avenue~\cite{Lu2013Abnormal}, UCSD Ped2~\cite{Li2014Anomaly}, UCF-Crime~\cite{Sultani2018Real}, ShanghaiTech~\cite{Luo2017A} and LAD databases.}
		\begin{tabular}{  c c   c  c  c  c  c }
			\toprule  
			Split&Subset& Avenue~\cite{Lu2013Abnormal}&UCSD Ped2~\cite{Li2014Anomaly}& ShanghaiTech~\cite{Luo2017A}&UCF-Crime~\cite{Sultani2018Real}&LAD\\
			\midrule  
			\multirow{2}{*}{Unsupervised}&   Train & 8   & 8  &175 & 800  &  958\\
			&   Test  & 18    & 14  & 199 & 290  &  560 \\
			\multirow{2}{*}{Weakly-supervised}&Train & 19   &  14 & 238 &1610 &  1440 \\
			&Test& 18  & 14  & 199 & 290   &  560 \\
			\multirow{2}{*}{Fully-supervised} &Train& 19  & 14  & 238 & 1610  &  1440 \\
			&Test & 18  & 14  & 199 &  290  &  560 \\
			\bottomrule  
		\end{tabular}	
		\label{tab:data_splits}
	\end{table*}

	\subsection{Implementation and Evaluation Metrics}\label{sec_experimetal_setup}
	
	\noindent{\textbf{Implementation}}: In this work, the proposed deep anomaly detection network is implemented in Ubuntu operating system with Tensorflow~\cite{65Abadi2016}. The experiments are conducted with Intel Core I7-6900K*16 CPU (3.20GHz), 64 GB RAM, and Nvidia TITAN X (Pascal) GPU with 16 GB memory.

	The I3D is pretrained by using Kinetics-400~\cite{CarreiraZ17}, which is a large-scale video classification dataset. We set the ${\lambda}_{1}$ and ${\lambda}_{2}$ as 1 and 10, respectively. We use a threshold of 0.5 to obtain the binarization anomaly score. Here, we use Adam optimizer~\cite{Kingma2014} to update parameters in the proposed model, the learning rate is set as 3e-4, the weight decay is set as 5e-4, the batch size is set as 60. 
	
	We divide the video sequence into clips with $m$=16 consecutive non-overlap video frames and set $K$=5. A total of $K$ $\times$ $m$=80 frames are used as the input to I3D to extract local spatiotemporal contextual features. The output of the sub-network of event classification is set as $C$=14, which is the number of LAD anomaly categories. We extract a $p$=1024 dimension feature of last pooling layer in the I3D, and concatenate the outputs of RGB and optical-flow I3D as the local spatiotemporal contextual feature of a video clip. The video frames are resized to $224 \times 224$, with their mean values being removed.
	
	The channel of hidden layers is set as 128 in proposed two-layer ConvLSTM, with $3 \times 3$ convolutional kernel and $1 \times 1$ stride. The dimension of the output features in our ConvLSTM is $4 \times 4 \times 128$. After reshaping the learned global spatiotemporal contextual feature as a 2048-dimension vector, we feed it into four fully convolutional layers for the final anomaly score prediction. The dimensions of the first three fully convolutional layers are set as 2048, 1024, and 512, respectively. The last layers are set as $C$=14 and $m \times K$=80 for anomalous categories classification and score prediction, respectively.
	
	\begin{table*}[ht]
		
		\centering 
		\renewcommand\tabcolsep{8pt} 
		\caption{AUC results on Avenue~\cite{Lu2013Abnormal}, UCSD Ped2~\cite{Li2014Anomaly}, UCF-Crime~\cite{Sultani2018Real}, ShanghaiTech~\cite{Luo2017A} and LAD databases, where $\mathcal{U}$, $\mathcal{W}$ and $\mathcal{S}$ represent unsupervised, weakly-supervised and fully-supervised methods, respectively. The * indicates experimental results are performed by using public source code.}
		\begin{tabular}{  c c   c  c  c  c  c }
			\toprule  
			\quad&\quad&                         Avenue~\cite{Lu2013Abnormal}&UCSD Ped2~\cite{Li2014Anomaly}& ShanghaiTech~\cite{Luo2017A}&UCF-Crime~\cite{Sultani2018Real}&LAD\\
			\midrule  
			Sparse~\cite{Lu2013Abnormal}&   $\mathcal{U}$	  & -  &  -  & -    & 65.51\,\;  &  50.31* \\
			ConvAE~\cite{Hasan2016Learning}&   $\mathcal{U}$  & -  &  -  & - & 50.60\,\;  &  53.24* \\
			GMM~\cite{Leyva2017Video} &     $\mathcal{U}$ 	  & - &  - & -     &  56.43* &  41.02* \\
			Stacked RNN~\cite{Luo2017A}&     $\mathcal{U}$    & 70.09* & 52.58*  & 67.66*  &-  & 49.42*         \\
			U-Net~\cite{Liu2018GAN} &      $\mathcal{U}$     & 55.26*  &  71.26*  & 56.59* &  - &  53.96* \\
			MNAD~\cite{park2020learning} &      $\mathcal{U}$     & 73.58*  &  46.72*  & 51.13* &  56.20* &  45.84* \\
			OGNet~\cite{zaheer2020old} &      $\mathcal{U}$     & 63.23*  &  69.08*  & 69.26* &  - &  55.07* \\
			
			\midrule  
			DeepMIL~\cite{Sultani2018Real}&   $\mathcal{W}$   & 87.53* &  90.19* & 86.30\,\; & \textbf{75.41}\,\;  &  70.18* \\
			MLEP~\cite{Liu2019MarginLE} &     $\mathcal{W}$   & 89.20\,\;  &-  & 73.40\,\; & 50.01*      & 50.57* \\
			
			AR-Net~\cite{wan2020weakly} &   $\mathcal{W}$    & 89.31* &    93.64* & 91.24\,\; &  74.36* &  79.84* \\
			\midrule  
			\textbf{Our Method} &        $\mathcal{F}$       & \textbf{89.33\,\;} & \textbf{95.12\,\;} & \textbf{92.97} & 74.98\,\; & \textbf{86.28\,\;} \\
			\bottomrule  
		\end{tabular}	
		\label{tab:AUC_results}
	\end{table*}
	
	\noindent{\textbf{Evaluation Metrics}}: In this study, following existing anomaly detection studies~\cite{Luo2017A, Liu2019MarginLE}, we utilize a frame-level Area Under ROC (Receiver Operating Characteristic) Curve (AUC) for quantitative performance evaluation. A higher AUC value indicates better performance. To evaluate anomalous categories performance of our model, we use the accuracy as the metric.

	\noindent{\textbf{Data Splits}}:We conduct the experiments upon six databases including Avenue~\cite{Lu2013Abnormal}, UCSD Ped2~\cite{Li2014Anomaly}, ShanghaiTech~\cite{Luo2017A}, UCF-Crime~\cite{Sultani2018Real} and LAD.
	We adopt three data splits of each database to fit requirements of unsupervised, weakly-supervised and fully-supervised anomaly detection methods.

	\noindent\textbf{Weakly-supervised Splits} In the standard protocol of Avenue, UCSD Ped2, ShanghaiTech, all training videos are normal, and this sitting is not suit for weakly-supervised learning. So we reorganize these databases. For Avenue, UCSD Ped2, we selected randomly 50\% video to be training video, while the rest are used as the testing set. We use the same splits from~\cite{wan2020weakly} and ~\cite{Sultani2018Real} for ShanghaiTech and UCF-Crime, respectively. Only video-level anomalous label is provided for training weakly-supervised anomaly detection models.
	
	\noindent\textbf{Unsupervised Splits} For each database, we use only normal videos in training set of the weakly-supervised split to train unsupervised anomaly detection models, and evaluate the unsupervised anomaly detection models by using all videos in test set of the weakly-supervised split. 
		
	\noindent\textbf{Fully-supervised Splits} We use the same data splits of weakly-supervised methods. Frame-level anomalous label and video-level anomalous categories label are provided for training model. It worthy notice that UCF-Crime does not provide frame-level anomalous label for the default training set, we use video-level anomalous label to be frame-level anomalous label for each training video.
	
	The numbers of training and testing videos on different splits of the databases are shown in Table~\ref{tab:data_splits}.
	
	\subsection{Comparison with State-of-the-art Models}\label{sec_evaluation_method}
	
	In this section, we compare the proposed approach with several state-of-the-art unsupervised and weakly-superivsed anomaly detection methods. The unsupervised anomaly detection methods contain GMM~\cite{Leyva2017Video}, Sparse~\cite{Lu2013Abnormal}, ConvAE~\cite{Hasan2016Learning}, Stack RNN~\cite{Luo2017A}, U-Net~\cite{Liu2018GAN}, MNAD~\cite{park2020learning} and OGNet~\cite{zaheer2020old}. The weakly-superivsed anomaly detection methods 
	contain DeepMIL~\cite{Sultani2018Real}, MLEP~\cite{Liu2019MarginLE} and AR-Net~\cite{wan2020weakly}. 
	
	GMM is an anomaly detection model by using Gaussian Mixture Model and Markov Chains. Sparse is a dictionary-based anomaly detection model by learning normal dictionary using sparse representation. ConvAE is the first deep learning based anomaly detection model by using Auto-Encoder to model normal event patterns. Stack RNN U-Net is an anomaly detection model based on the reconstruction errors between a predicted frame and the ground-truth. MNAD and OGNet are latest unsupervised anomaly detection methods. We retrain these models by using unsupervised splits on each database in this paper. DeepMIL is a representative weakly-supervised anomaly detection method, and AR-Net achieves the highest AUC performance so far in ShanghaiTech. The performance of weakly-supervised anomaly detection methods are obtained by using weakly-supervised splits on each database. It should notices that unsupervised anomaly detection methods only use the normal videos to train their models.

	We show the comparison results in Table~\ref{tab:AUC_results} on Avenue, UCSD Ped2, ShanghaiTech, UCF-Crime and LAD. Our method outperforms all competing anomaly detection models on our LAD and achieves an absolute gain of 6.44\% in terms of AUC, compared to the state-of-the-art~\cite{wan2020weakly}. Compare the weakly-supervised methods, our method achieves similar AUC performance on UCF-Crime. It may be caused by using noisy frame-level anomalous label, which is the same as video-level anomalous label, to train our model. Our model achieves higher AUC performance than the competing anomaly detection models on Avenue, UCSD Ped2 and ShanghaiTech, it reveals frame-level annotation is an effective tool to promote anomaly detection task.
	
	As shown in Table~\ref{tab:AUC_results}, the competing weakly-supervised anomaly detection models obtain higher AUC performance on Avenue, UCSD Ped and  ShanghaiTech, compare to unsupervised anomaly detection models. It indicates that the database including training abnormal videos is necessity to promote the video anomaly detection task. Most competing models achieve higher AUC performance on Avenue and UCSD Ped, while they get relative lower AUC performance on ShanghaiTech, UCF-Crime and LAD. These experimental results demonstrate that variety of visual scenes is a indeterminable issue to current anomaly detection models. It indicates that our LAD, which contains thousands visual scenes, is a challenge database for video anomaly detection.

	\begin{table}[ht]
		\centering 
		\renewcommand\tabcolsep{18pt} 
		\caption{Competition of different local spatiotemporal feature extractor, where \textbf{R}\\ and \textbf{O} indicate RGB and Optical-flow, respectively.}
		\begin{tabular}{ c  c  c}
			\toprule  
			\quad& Input modal & AUC\\
			\midrule  
			C3D~\cite{Du2015Learning} & \textbf{R} & 77.21 \\
			$\rm I3D^{RGB}$~\cite{CarreiraZ17} & \textbf{R} & 84.43 \\
			$\rm I3D^{Optical-flow}$~\cite{CarreiraZ17} & \textbf{O} & 82.46 \\
			I3D~\cite{CarreiraZ17} & \textbf{R}{\&}\textbf{O} & \textbf{86.93}\\
			\bottomrule  
		\end{tabular}
		\label{tab:ablation_feature}
	\end{table}

	\subsection{Ablation Study}\label{Ablation}
	To evaluate the effectiveness of the local spatiotemporal feature extractor, we compare four different spatiotemporal networks including C3D~\cite{Du2015Learning}, $\rm I3D^{RGB}$~\cite{CarreiraZ17}, $\rm I3D^{Optical-flow}$~\cite{CarreiraZ17} and I3D~\cite{CarreiraZ17} on LAD. As shown in Table~\ref{tab:ablation_feature}, our method with C3D achieves a frame-level AUC of 77.21\%. And the $\rm I3D^{RGB}$ and $\rm I3D^{RGB}$ based our method achieves 84.43\% and 82.46\% in terms of AUC, respectively. Our Method with I3D boosts the performance with a frame-level AUC of 86.93\%.
	
	The comparison results by using different loss functions in Table~\ref{tab:lossfunction} illustrate the boost brought by the proposed multi-task loss functions. Our method with $L_{1}$, a loss function for the anomaly score prediction task, is treated as the baseline. It achieves a frame-level AUC of 80.43\% on LAD. While our method with both $L_{1}$ and $L_{2}$ obtains an absolute gain of 6.50\% in terms of AUC. And the $L_{1}$ loss function boosts the anomalous categories performance by obtaining a 59.3\% in terms of accuracy.

	\begin{table}[ht]
		\centering 
		\renewcommand\tabcolsep{18pt} 
		\caption{AUC and accuracy results obtained by using different loss functions\\ on LAD.}
		\begin{tabular}{ c  c  c  c}
			\toprule  
			$L_{1}$ & $L_{2}$ & AUC & Accuracy\\
			\midrule  
			\Checkmark & \XSolid  & 80.43 &  3.1 \\
			\XSolid   & \Checkmark & 50.00  & 58.4 \\
			\Checkmark & \Checkmark & \textbf{86.93}  & \textbf{59.3} \\
			\bottomrule  
		\end{tabular}
		
		\label{tab:lossfunction}
	\end{table}

	\subsection{Qualitative Analysis}\label{Qualitative Analysis}
	
	\begin{figure}[ht]
		\setlength{\abovecaptionskip}{2pt}
		\centering
		\includegraphics[width=0.48\textwidth]{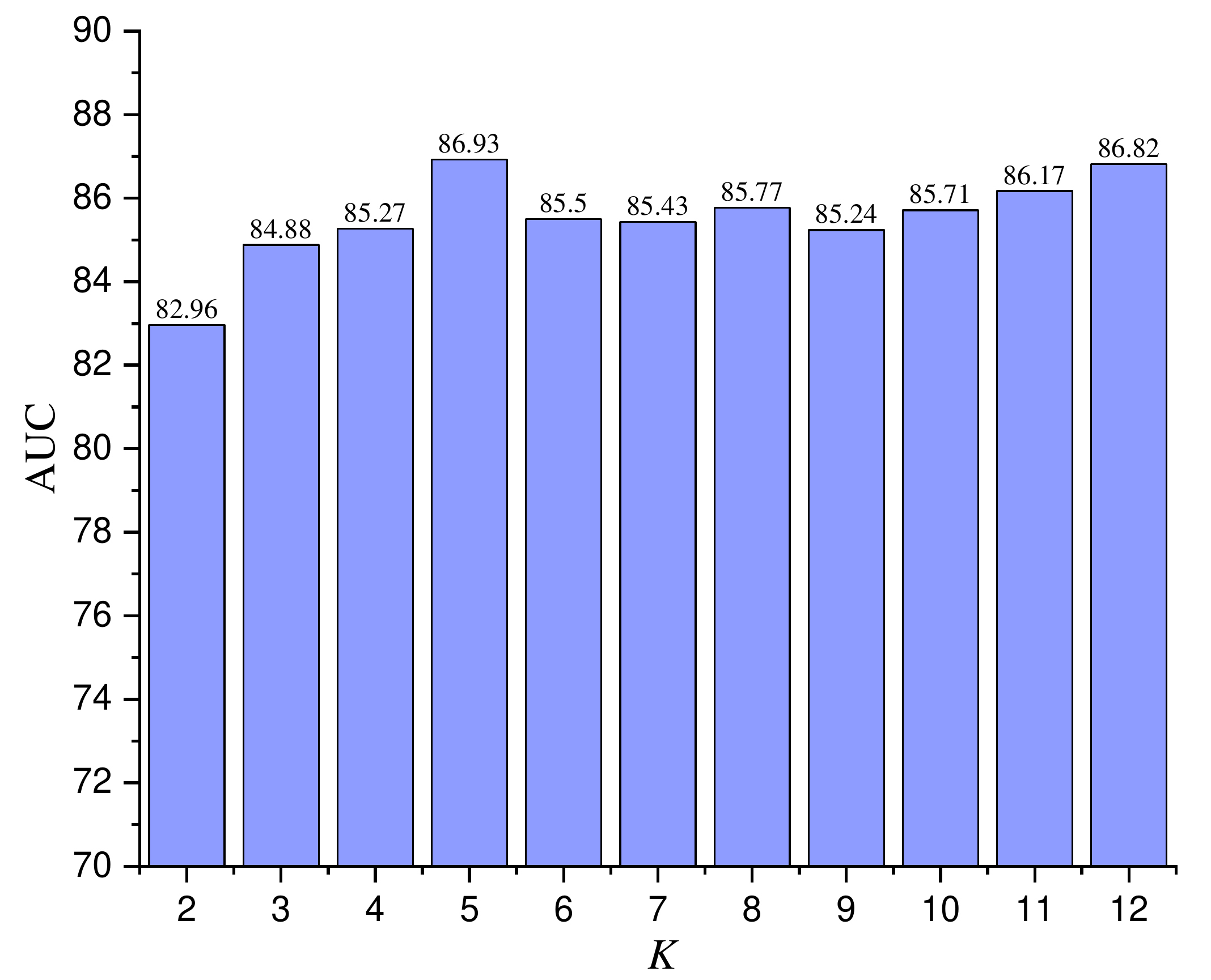}
		\caption{AUC of different $K$ values.}\label{fig:k}
	\end{figure}
	
\begin{figure}[ht]
		\setlength{\abovecaptionskip}{2pt}
		\centering
		\includegraphics[width=0.48\textwidth]{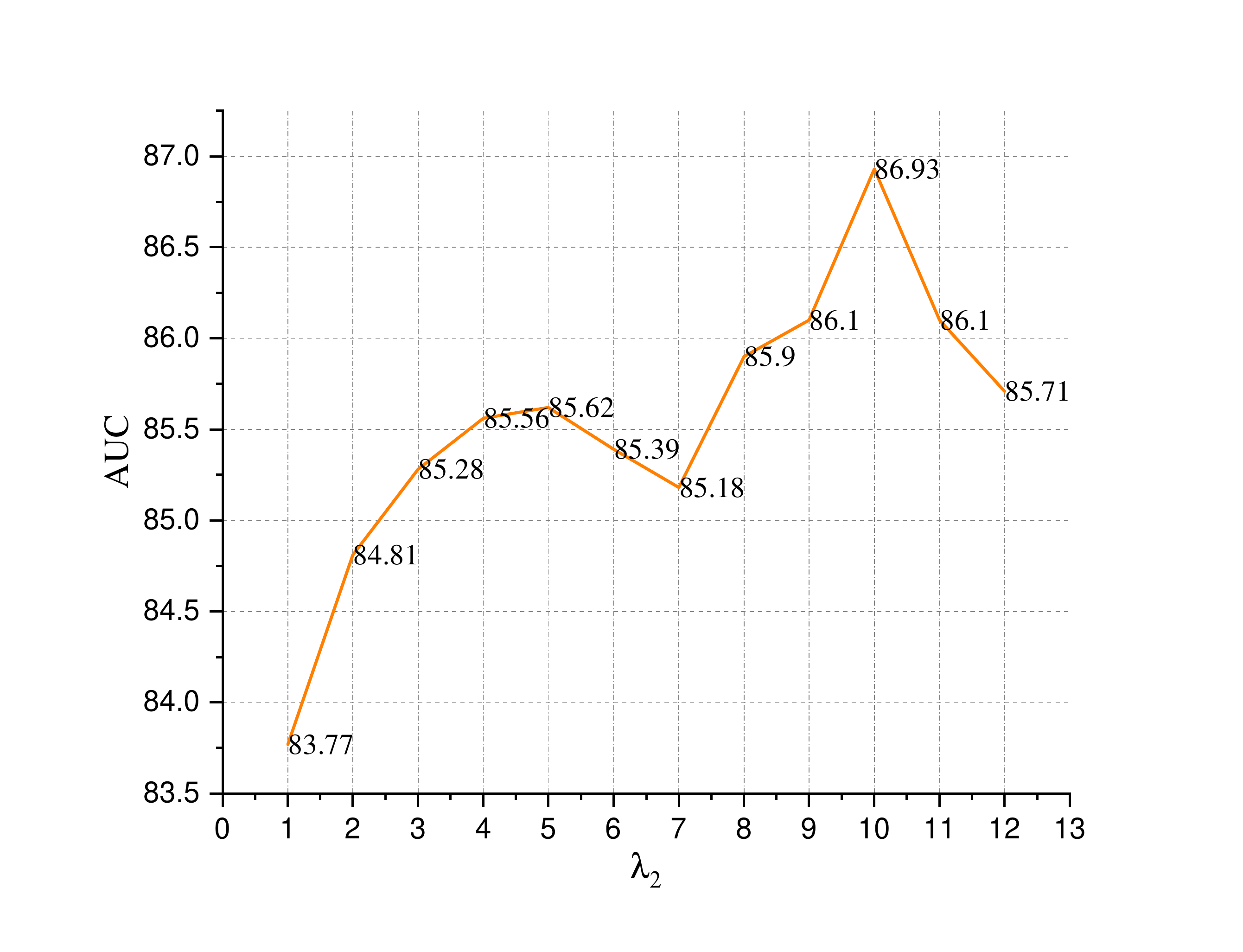}
		\caption{AUC of different $\lambda_{2}$ values.}
		\label{fig:lossw}
	\end{figure}

	In order to gain insight into the hyperparameter $K$, we perform experiments using the I3D local spatiotemporal feature extractor with different values of $K$, as shown in Fig.~\ref{fig:k}. Our method achieves the best performance in terms of AUC when we set $K=5$, and our method with  $K=2$ obtains an absolute reduction of 3.97\% in terms of AUC. It confirms the necessity to gain global spatiotemporal features. The comparison results by using different $\lambda_{2}$ values are shown in Fig.~\ref{fig:lossw}, and we set $\lambda_{1}$=1 in these experiments. Our method gains the boost when $\lambda_{2}$ is set as 10.

	\begin{figure}[ht]
		\setlength{\abovecaptionskip}{2pt}
		\centering
		\includegraphics[width=0.5\textwidth]{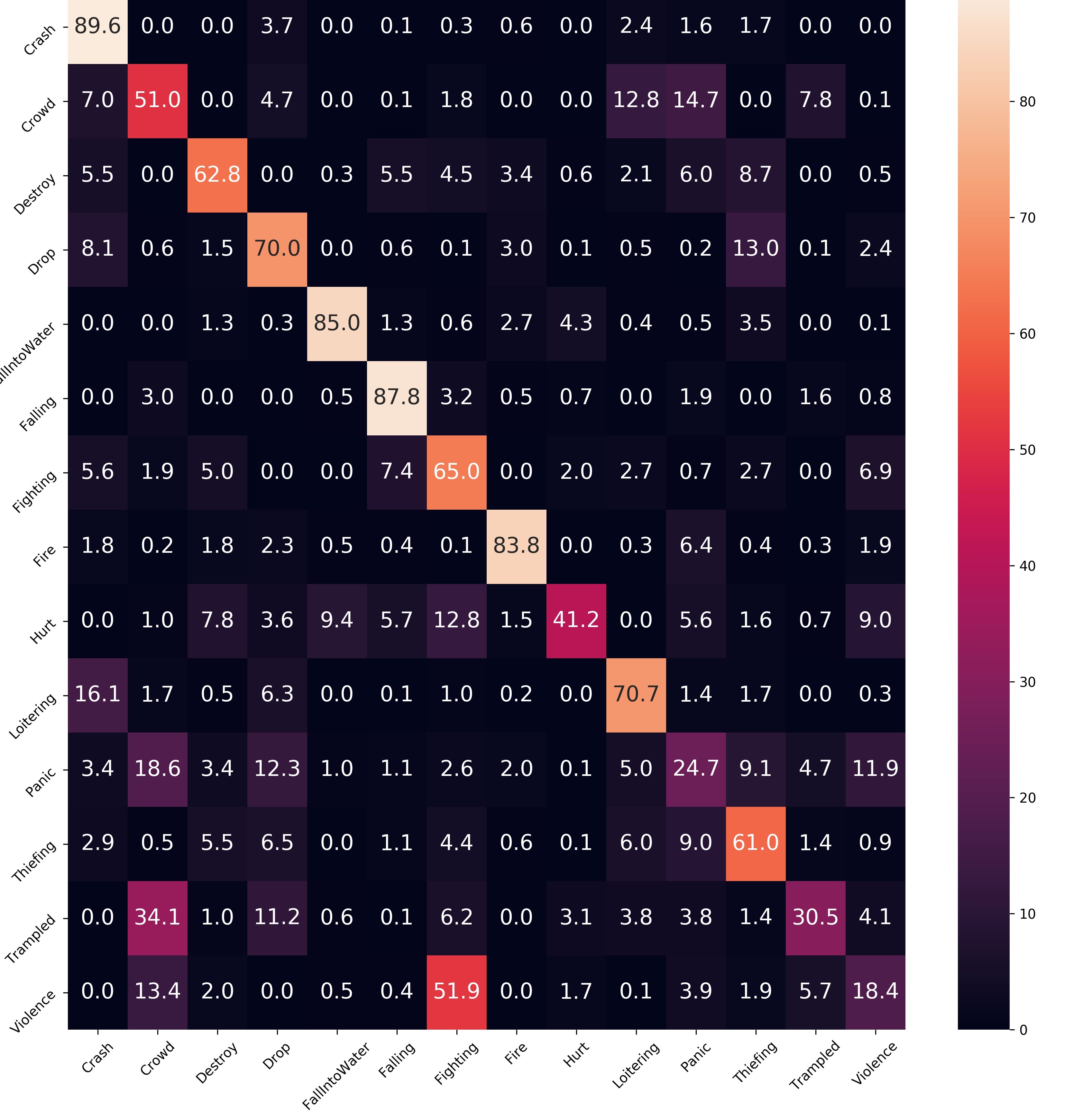}
		\caption{Visualization of the confusion matrix of anomalous categories classification results by using our method.
		}\label{fig:Confusion}
	\end{figure}

		\begin{table}[ht]
		\centering 
		\renewcommand\tabcolsep{18pt} 
		\caption{Experimental results of anomalous categories classification.}
		\begin{tabular}{  c  c  c }
			\toprule  
			\quad & UCF-Crime &LAD\\
			\midrule  
			TCNN~\cite{hou2017tube} & 28.4 & - \\
			C3D~\cite{Du2015Learning} & 23.0 & 45.9 \\
			Our Method & \textbf{59.6} & \textbf{59.3} \\
			\bottomrule  
		\end{tabular}
		
		\label{tab:Time_and_accuracy}
	\end{table}

	As shown in Table \ref{tab:Time_and_accuracy}, we can observe that our method outperforms the competing model on UCF-Crime and LAD. Specifically, Our method obtains a relative gain of over 100\% in terms of accuracy on UCF-Crime, compared to TCNN~\cite{hou2017tube} and C3D. Comparing to C3D, our method boosts the accuracy performance by obtaining a 13.4\% absolute improvement in terms of accuracy.
	
	To analyze the anomalous categories classification performance, we show the confusion matrix of our method in Fig.~\ref{fig:Confusion}, where we can observe that the proposed model can obtain a promising performance of abnormal event classification. The worst accuracy is from the violence category since the anomaly samples of the violence category are easily wrongly classified into the crowd or fighting categories. The best accuracy is obtained from the crash, falling, fire and fallingtowater categories since their anomaly definitions are clear, and there is a big gap between these categories and other categories.

	\section{Conclusion}\label{sec_conclusion}
	In this study, we contribute a large-scale benchmark for anomaly detection in video sequences. It contains 2000 different video sequences with 14 anomaly categories. We provide annotation data including video-level and frame-level labels. The proposed database enables research possibility of anomaly detection in a fully-supervised manner. Then we propose a multi-task computational model of anomaly detection by effectively learning local and global spatiotemporal contextual features for video sequences. In the proposed multi-task deep neural network, the local spatiotemporal features are first extracted by an Inflated $3$D convolutional network from each video segment. Then we feed these local spatiotemporal contextual features to a recurrent convolutional architecture to learn global spatiotemporal contextual features. Finally, anomaly scores and abnormal event categories are predicted by the output of the fully convolutional layers of two sub-networks. Comparison experiments show that the proposed method outperforms the state-of-the-art anomaly detection methods on public databases and the built LAD database. In the future, we will further investigate anomaly detection to improve the performance of anomaly detection for video sequences.
	
	\section{Acknowledgment}\label{Acknowledgment}
	This work was supported in part by the Jiangxi Provincial Natural Science Foundation under 20202ACB202007, the Fok Ying Tung Education Foundation under 161061, the Foundation of Jiangxi Provincial Department of Education under Grants 20203BBE53033, the Youth Foundation of Jiangxi Education Department under Grants GJJ200535 and GJJ190279, and the Postgraduate Innovation Special Fund of Jiangxi Province, China under Grant YC2020-B139.
	
	\bibliographystyle{IEEEtran}
	\bibliography{wanbib1}

\begin{thebibliography}{10}
\providecommand{\url}[1]{#1}
\csname url@samestyle\endcsname
\providecommand{\newblock}{\relax}
\providecommand{\bibinfo}[2]{#2}
\providecommand{\BIBentrySTDinterwordspacing}{\spaceskip=0pt\relax}
\providecommand{\BIBentryALTinterwordstretchfactor}{4}
\providecommand{\BIBentryALTinterwordspacing}{\spaceskip=\fontdimen2\font plus
\BIBentryALTinterwordstretchfactor\fontdimen3\font minus
  \fontdimen4\font\relax}
\providecommand{\BIBforeignlanguage}[2]{{%
\expandafter\ifx\csname l@#1\endcsname\relax
\typeout{** WARNING: IEEEtran.bst: No hyphenation pattern has been}%
\typeout{** loaded for the language `#1'. Using the pattern for}%
\typeout{** the default language instead.}%
\else
\language=\csname l@#1\endcsname
\fi
#2}}
\providecommand{\BIBdecl}{\relax}
\BIBdecl

\bibitem{Li2014Anomaly}
W.~Li, V.~Mahadevan, and N.~Vasconcelos, ``Anomaly detection and localization
  in crowded scenes,'' \emph{IEEE Transactions on Pattern Analysis and Machine
  Intelligence}, vol.~36, no.~1, pp. 18--32, 2013.

\bibitem{Cosar2017Toward}
S.~Cosar, G.~Donatiello, V.~Bogorny, C.~Garate, L.~O. Alvares, and F.~Bremond,
  ``Toward abnormal trajectory and event detection in video surveillance,''
  \emph{IEEE Transactions on Circuits and Systems for Video Technology},
  vol.~27, no.~3, pp. 683--695, 2017.

\bibitem{Ravanbakhsh2017Abnormal}
M.~Ravanbakhsh, M.~Nabi, E.~Sangineto, L.~Marcenaro, C.~Regazzoni, and N.~Sebe,
  ``Abnormal event detection in videos using generative adversarial nets,'' in
  \emph{IEEE International Conference on Image Processing}, 2017, pp.
  1577--1581.

\bibitem{Hasan2016Learning}
M.~Hasan, J.~Choi, J.~Neumann, A.~K. Roy-Chowdhury, and L.~S. Davis, ``Learning
  temporal regularity in video sequences,'' in \emph{IEEE Conference on
  Computer Vision and Pattern Recognition}, 2016, pp. 733--742.

\bibitem{Liu2018GAN}
W.~Liu, W.~Luo, D.~Lian, and S.~Gao, ``Future frame prediction for anomaly
  detection--a new baseline,'' in \emph{IEEE Conference on Computer Vision and
  Pattern Recognition}, 2018, pp. 6536--6545.

\bibitem{wan2020weakly}
B.~Wan, Y.~Fang, X.~Xia, and J.~Mei, ``Weakly supervised video anomaly
  detection via center-guided discriminative learning,'' in \emph{IEEE
  International Conference on Multimedia and Expo}, 2020, pp. 1--6.

\bibitem{park2020learning}
H.~Park, J.~Noh, and B.~Ham, ``Learning memory-guided normality for anomaly
  detection,'' in \emph{IEEE Conference on Computer Vision and Pattern
  Recognition}, 2020, pp. 14\,372--14\,381.

\bibitem{zaheer2020old}
M.~Z. Zaheer, J.-h. Lee, M.~Astrid, and S.-I. Lee, ``Old is gold: Redefining
  the adversarially learned one-class classifier training paradigm,'' in
  \emph{IEEE Conference on Computer Vision and Pattern Recognition}, 2020, pp.
  14\,183--14\,193.

\bibitem{Liu2019MarginLE}
W.~Liu, W.~Luo, Z.~Li, P.~Zhao, and S.~Gao, ``Margin learning embedded
  prediction for video anomaly detection with a few anomalies,'' in
  \emph{International Joint Conference on Artificial Intelligence}, 2019, pp.
  3023--3030.

\bibitem{Antic2011Video}
B.~Anti{\'c} and B.~Ommer, ``Video parsing for abnormality detection,'' in
  \emph{IEEE Conference on Computer Vision and Pattern Recognition}, 2011, pp.
  2415--2422.

\bibitem{Lu2013Abnormal}
C.~Lu, J.~Shi, and J.~Jia, ``Abnormal event detection at 150 fps in matlab,''
  in \emph{IEEE Conference on Computer Vision and Pattern Recognition}, 2013,
  pp. 2720--2727.

\bibitem{Leyva2017Video}
R.~Leyva, V.~Sanchez, and C.-T. Li, ``Video anomaly detection with compact
  feature sets for online performance,'' \emph{IEEE Transactions on Image
  Processing}, vol.~26, no.~7, pp. 3463--3478, 2017.

\bibitem{Luo2017A}
W.~Luo, L.~Wen, and S.~Gao, ``A revisit of sparse coding based anomaly
  detection in stacked rnn framework,'' in \emph{IEEE International Conference
  on Computer Vision}, 2017, pp. 341--349.

\bibitem{Sultani2018Real}
W.~Sultani, C.~Chen, and M.~Shah, ``Real-world anomaly detection in
  surveillance videos,'' in \emph{IEEE Conference on Computer Vision and
  Pattern Recognition}, 2018, pp. 6479--6488.

\bibitem{Zhu2013Anomaly}
X.~Zhu, J.~Liu, J.~Wang, Y.~Fang, and H.~Lu, ``Anomaly detection in crowded
  scene via appearance and dynamics joint modeling,'' in \emph{IEEE
  International Conference on Image Processing}, 2013, pp. 2705--2708.

\bibitem{Colque2015Histograms}
R.~V. H.~M. Colque, C.~A.~C. Júnior, and W.~R. Schwartz, ``Histograms of
  optical flow orientation and magnitude to detect anomalous events in
  videos,'' in \emph{Sibgrapi Conference on Graphics, Patterns and Images},
  2015, pp. 126--133.

\bibitem{yuan2015hyperspectral}
Y.~Yuan, D.~Ma, and Q.~Wang, ``Hyperspectral anomaly detection by graph pixel
  selection,'' \emph{IEEE Transactions on Cybernetics}, vol.~46, no.~12, pp.
  3123--3134, 2015.

\bibitem{Mohammadi2016Angry}
S.~Mohammadi, A.~Perina, H.~Kiani, and V.~Murino, ``Angry crowds: Detecting
  violent events in videos,'' in \emph{European Conference on Computer Vision},
  2016, pp. 3--18.

\bibitem{Sultani2010Abnormal}
W.~Sultani and Y.~C. Jin, ``Abnormal traffic detection using intelligent driver
  model,'' in \emph{International Conference on Pattern Recognition}, 2010, pp.
  324--327.

\bibitem{yuan2016anomaly}
Y.~Yuan, D.~Wang, and Q.~Wang, ``Anomaly detection in traffic scenes via
  spatial-aware motion reconstruction,'' \emph{IEEE Transactions on Intelligent
  Transportation Systems}, vol.~18, no.~5, pp. 1198--1209, 2016.

\bibitem{Piciarelli2008Trajectory}
C.~Piciarelli, C.~Micheloni, and G.~L. Foresti, ``Trajectory-based anomalous
  event detection,'' \emph{IEEE Transactions on Circuits and Systems for Video
  Technology}, vol.~18, no.~11, pp. 1544--1554, 2008.

\bibitem{Wu2010Chaotic}
S.~Wu, B.~E. Moore, and M.~Shah, ``Chaotic invariants of lagrangian particle
  trajectories for anomaly detection in crowded scenes,'' in \emph{IEEE
  Conference on Computer Vision and Pattern Recognition}, 2010, pp. 2054--2060.

\bibitem{piciarelli2006on-line}
C.~Piciarelli and G.~L. Foresti, ``On-line trajectory clustering for anomalous
  events detection,'' \emph{Pattern Recognition Letters}, vol.~27, no.~15, pp.
  1835--1842, 2006.

\bibitem{jiang2011anomalous}
F.~Jiang, J.~Yuan, S.~A. Tsaftaris, and A.~K. Katsaggelos, ``Anomalous video
  event detection using spatiotemporal context,'' \emph{Computer Vision and
  Image Understanding}, vol. 115, no.~3, pp. 323--333, 2011.

\bibitem{tung2011goal-based}
F.~Tung, J.~S. Zelek, and D.~A. Clausi, ``Goal-based trajectory analysis for
  unusual behaviour detection in intelligent surveillance,'' \emph{Image and
  Vision Computing}, vol.~29, no.~4, pp. 230--240, 2011.

\bibitem{morris2011trajectory}
B.~Morris and M.~M. Trivedi, ``Trajectory learning for activity understanding:
  Unsupervised, multilevel, and long-term adaptive approach,'' \emph{IEEE
  Transactions on Pattern Analysis and Machine Intelligence}, vol.~33, no.~11,
  pp. 2287--2301, 2011.

\bibitem{calderara2011detecting}
S.~Calderara, U.~Heinemann, A.~Prati, R.~Cucchiara, and N.~Tishby, ``Detecting
  anomalies in people's trajectories using spectral graph analysis,''
  \emph{Computer Vision and Image Understanding}, vol. 115, no.~8, pp.
  1099--1111, 2011.

\bibitem{patino2015abnormal}
L.~Patino, J.~Ferryman, and C.~Beleznai, ``Abnormal behaviour detection on
  queue analysis from stereo cameras,'' in \emph{IEEE International Conference
  on Advanced Video and Signal Based Surveillance}, 2015, pp. 1--6.

\bibitem{yi2015understanding}
S.~Yi, H.~Li, and X.~Wang, ``Understanding pedestrian behaviors from stationary
  crowd groups,'' in \emph{IEEE Conference on Computer Vision and Pattern
  Recognition}, 2015, pp. 3488--3496.

\bibitem{Wang2014Detection}
T.~Wang and H.~Snoussi, ``Detection of abnormal visual events via global
  optical flow orientation histogram,'' \emph{IEEE Transactions on Information
  Forensics and Security}, vol.~9, no.~6, pp. 988--998, 2014.

\bibitem{Mehran2009Abnormal}
R.~Mehran, A.~Oyama, and M.~Shah, ``Abnormal crowd behavior detection using
  social force model,'' in \emph{IEEE Conference on Computer Vision and Pattern
  Recognition}, 2009, pp. 935--942.

\bibitem{adam2008robust}
A.~Adam, E.~Rivlin, I.~Shimshoni, and D.~Reinitz, ``Robust real-time unusual
  event detection using multiple fixed-location monitors,'' \emph{IEEE
  Transactions on Pattern Analysis and Machine Intelligence}, vol.~30, no.~3,
  pp. 555--560, 2008.

\bibitem{saligrama2012video}
V.~Saligrama and Z.~Chen, ``Video anomaly detection based on local statistical
  aggregates,'' in \emph{IEEE Conference on Computer Vision and Pattern
  Recognition}, 2012, pp. 2112--2119.

\bibitem{benezeth2009abnormal}
Y.~Benezeth, P.~M. Jodoin, V.~Saligrama, and C.~Rosenberger, ``Abnormal events
  detection based on spatio-temporal co-occurences,'' in \emph{IEEE Conference
  on Computer Vision and Pattern Recognition}, 2012, pp. 2458--2465.

\bibitem{kim2009observe}
J.~Kim and K.~Grauman, ``Observe locally, infer globally: a space-time mrf for
  detecting abnormal activities with incremental updates,'' in \emph{IEEE
  Conference on Computer Vision and Pattern Recognition}, 2009, pp. 2921--2928.

\bibitem{kratz2009anomaly}
L.~Kratz and K.~Nishino, ``Anomaly detection in extremely crowded scenes using
  spatio-temporal motion pattern models,'' in \emph{IEEE Conference on Computer
  Vision and Pattern Recognition}, 2009, pp. 1446--1453.

\bibitem{zhang2005semi-supervised}
D.~Zhang, D.~Gatica-Perez, S.~Bengio, and I.~A. Mccowan, ``Semi-supervised
  adapted hmms for unusual event detection,'' in \emph{IEEE Computer Society
  Conference on Computer Vision and Pattern Recognition}, 2005, pp. 611--618.

\bibitem{roshtkhari2013online}
M.~J. Roshtkhari and M.~D. Levine, ``Online dominant and anomalous behavior
  detection in videos,'' in \emph{IEEE Conference on Computer Vision and
  Pattern Recognition}, 2013, pp. 2611--2618.

\bibitem{zhu2013context-aware}
Y.~Zhu, N.~M. Nayak, and A.~K. Roy-Chowdhury, ``Context-aware modeling and
  recognition of activities in video,'' in \emph{IEEE Conference on Computer
  Vision and Pattern Recognition}, 2013, pp. 2491--2398.

\bibitem{xiao2015learning}
T.~Xiao, C.~Zhang, and H.~Zha, ``Learning to detect anomalies in surveillance
  video,'' \emph{IEEE Signal Processing Letters}, vol.~22, no.~9, pp.
  1477--1481, 2015.

\bibitem{Cui2011Abnormal}
X.~Cui, Q.~Liu, M.~Gao, and D.~N. Metaxas, ``Abnormal detection using
  interaction energy potentials,'' in \emph{IEEE Conference on Computer Vision
  and Pattern Recognition}, 2011, pp. 3161--3167.

\bibitem{Yuan2015Online}
Y.~Yuan, J.~Fang, and Q.~Wang, ``Online anomaly detection in crowd scenes via
  structure analysis,'' \emph{IEEE Transactions on Cybernetics}, vol.~45,
  no.~3, pp. 548--561, 2015.

\bibitem{Cheng2015Gaussian}
Y.~Zhang, L.~Qin, R.~Ji, H.~Yao, and Q.~Huang, ``Gaussian process
  regression-based video anomaly detection and localization with hierarchical
  feature representation,'' \emph{IEEE Transactions on Image Processing},
  vol.~24, no.~12, pp. 5288--5301, 2015.

\bibitem{athanesious2020detecting}
J.~Athanesious, V.~Srinivasan, V.~Vijayakumar, S.~Christobel, and S.~C.
  Sethuraman, ``Detecting abnormal events in traffic video surveillance using
  superorientation optical flow feature,'' \emph{IET Image Processing},
  vol.~14, no.~9, pp. 1881--1891, 2020.

\bibitem{Sabokrou2017Deep}
M.~Sabokrou, M.~Fayyaz, M.~Fathy, and R.~Klette, ``Deep-cascade: Cascading 3d
  deep neural networks for fast anomaly detection and localization in crowded
  scenes,'' \emph{IEEE Transactions on Image Processing}, vol.~26, no.~4, pp.
  1992--2004, 2017.

\bibitem{sabokrou2016video}
M.~Sabokrou, M.~Fathy, and M.~Hoseini, ``Video anomaly detection and
  localisation based on the sparsity and reconstruction error of
  auto-encoder,'' \emph{Electronics Letters}, vol.~52, no.~13, pp. 1122--1124,
  2016.

\bibitem{Luo2017Remembering}
W.~Luo, L.~Wen, and S.~Gao, ``Remembering history with convolutional lstm for
  anomaly detection,'' in \emph{IEEE International Conference on Multimedia and
  Expo}, 2017, pp. 439--444.

\bibitem{Hinami2017Joint}
R.~Hinami, M.~Tao, and S.~Satoh, ``Joint detection and recounting of abnormal
  events by learning deep generic knowledge,'' in \emph{IEEE International
  Conference on Computer Vision}, 2017, pp. 3639--3647.

\bibitem{Ionescu2017Unmasking}
R.~Tudor~Ionescu, S.~Smeureanu, B.~Alexe, and M.~Popescu, ``Unmasking the
  abnormal events in video,'' in \emph{IEEE International Conference on
  Computer Vision}, 2017, pp. 2895--2903.

\bibitem{Xu2015learning}
X.~Dan, R.~Elisa, Y.~Yan, S.~Jingkuan, and S.~Nicu, ``Learning deep
  representations of appearance and motion for anomalous event detection,'' in
  \emph{British Machine Vision Conference}, 2015, pp. 1--12.

\bibitem{FCVID}
Y.~Jiang, Z.~Wu, J.~Wang, X.~Xue, and S.~Chang, ``Exploiting feature and class
  relationships in video categorization with regularized deep neural
  networks,'' \emph{IEEE Transactions on Pattern Analysis and Machine
  Intelligence}, vol.~40, no.~2, pp. 352--364, 2018.

\bibitem{Marszalek2009Actions}
M.~Marszalek, I.~Laptev, and C.~Schmid, ``Actions in context,'' in \emph{IEEE
  Conference on Computer Vision and Pattern Recognition}, 2009, pp. 2929--2936.

\bibitem{Liu2009Recognizing}
J.~Liu, J.~Luo, and M.~Shah, ``Recognizing realistic actions from videos,'' in
  \emph{IEEE Conference on Computer Vision and Pattern Recognition}, 2009, pp.
  1996--2003.

\bibitem{CarreiraZ17}
J.~Carreira and A.~Zisserman, ``Quo vadis, action recognition? {A} new model
  and the kinetics dataset,'' in \emph{IEEE Conference on Computer Vision and
  Pattern Recognition}, 2017, pp. 4724--4733.

\bibitem{Shi2015Convolutional}
X.~Shi, Z.~Chen, H.~Wang, D.-Y. Yeung, W.-k. Wong, and W.-c. Woo,
  ``Convolutional lstm network: a machine learning approach for precipitation
  nowcasting,'' in \emph{International Conference on Neural Information
  Processing Systems}, 2015, pp. 802--810.

\bibitem{Girshick2015Fast}
R.~Girshick, ``Fast r-cnn,'' in \emph{IEEE International Conference on Computer
  Vision}, 2015, pp. 1440--1448.

\bibitem{65Abadi2016}
M.~Abadi, A.~Agarwal, P.~Barham, E.~Brevdo, Z.~Chen, C.~Citro, G.~S. Corrado,
  A.~Davis, J.~Dean, M.~Devin \emph{et~al.}, ``Tensorflow: Large-scale machine
  learning on heterogeneous distributed systems,'' \emph{arXiv preprint
  arXiv:1603.04467}, 2016.

\bibitem{Kingma2014}
D.~P. Kingma and J.~Ba, ``Adam: {A} method for stochastic optimization,'' in
  \emph{International Conference on Learning Representations}, 2015.

\bibitem{Du2015Learning}
T.~Du, L.~Bourdev, R.~Fergus, L.~Torresani, and M.~Paluri, ``Learning
  spatiotemporal features with 3d convolutional networks,'' in \emph{IEEE
  International Conference on Computer Vision}, 2015, pp. 4489--4497.

\bibitem{hou2017tube}
R.~Hou, C.~Chen, and M.~Shah, ``Tube convolutional neural network (t-cnn) for
  action detection in videos,'' in \emph{IEEE International Conference on
  Computer Vision}, 2017, pp. 5822--5831.

\end{thebibliography}
	
\end{document}